\documentclass[11pt,letterpaper]{article}

\usepackage[preprint]{acl}
\usepackage{times}
\usepackage{latexsym}
\usepackage[T1]{fontenc}
\usepackage[utf8]{inputenc}
\usepackage{microtype}

\usepackage{amsmath, amssymb, amsthm, bm}

\usepackage{booktabs, multirow, tabularx}
\usepackage{makecell}
\usepackage[table]{xcolor}
\usepackage{colortbl}

\usepackage{graphicx}
\usepackage{tikz}
\usetikzlibrary{positioning, arrows.meta, shapes.geometric, fit}

\usepackage{pifont}
\newcommand{\cmark}{\textcolor{green!55!black}{\ding{51}}}
\newcommand{\xmark}{\textcolor{red!75!black}{\ding{55}}}
\newcommand{\partialmark}{\textcolor{orange!85!black}{$\bullet$}}

\usepackage{url, hyperref}
\usepackage[capitalize,noabbrev]{cleveref}

\newcommand{\R}{\ensuremath{\mathbb{R}}}

\newcommand{\Acc}{\ensuremath{\mathrm{Acc}}}
\newcommand{\KL}{\ensuremath{\mathrm{KL}}}

\newcommand{\f}{\ensuremath{f_\theta}}                 
\newcommand{\hh}{\ensuremath{\bm{h}}}                  
\newcommand{\dhid}{\ensuremath{\Delta\hh}}             
\newcommand{\layer}{\ensuremath{\ell}}

\newcommand{\Vcs}{\ensuremath{V_{\mathrm{cs}}}}        
\newcommand{\Vns}{\ensuremath{V_{\mathrm{ns}}}}        
\newcommand{\Vrand}{\ensuremath{V_{\mathrm{rand}}}}    

\newcommand{\CLadder}{\textsc{CLadder}}
\newcommand{\CounterBench}{\textsc{CounterBench}}

\newcommand{\acs}{\ensuremath{\text{anti-CS}}}
\newcommand{\cs}{\ensuremath{\text{cs}}}
\newcommand{\ns}{\ensuremath{\text{ns}}}

\crefname{definition}{Definition}{Definitions}
\crefname{proposition}{Proposition}{Propositions}
\crefname{assumption}{Assumption}{Assumptions}

\definecolor{strongred}{HTML}{C0392B}
\definecolor{strongblue}{HTML}{2980B9}
\definecolor{paleCream}{HTML}{FAF6EE}   
\definecolor{warmAmber}{HTML}{F4D9A8}   
\definecolor{sageGrey} {HTML}{D8DDC9}   
\definecolor{deepSage} {HTML}{8A9A78}   
\definecolor{mechA}{HTML}{F4D9A8}  
\definecolor{mechB}{HTML}{FAF6EE}  
\definecolor{mechC}{HTML}{D8DDC9}  
\definecolor{mechD}{HTML}{F4D9A8}  
\definecolor{mechE}{HTML}{D8DDC9}  

\title{Causal Tongue-Tie: LLMs Can Encode Causal Direction, But Their Yes/No Outputs Fail to Express}

\author{%
  Ziyi Ding \\
  Tsinghua Shenzhen International\\Graduate School, Tsinghua University\\
  Shenzhen, China \\
  \And
  Xiao-Ping Zhang\thanks{Corresponding author: \texttt{xpzhang@ieee.org}} \\
  Tsinghua Shenzhen International\\Graduate School, Tsinghua University\\
  Shenzhen, China \\
}

\begin{document}
\maketitle

\begin{abstract}
We find a mismatch between what large language models encode about a
causal question and what they answer. On \emph{anti-commonsense}
\CLadder{}~\citep{jin2024cladder} items, a fixed linear probe recovers
the evidence-supported answer from the model's hidden state
(accuracy ${\approx}0.97$), while the spoken Yes/No reverts to the
commonsense one (accuracy ${\approx}0.5$). We call this
$\Delta {\approx} {+}0.5$ gap \emph{Causal Tongue-Tie}: a wrong Yes/No
decomposes into two separable failure modes---no internal signal vs.\
a signal the verbal interface cannot say. The implication cuts both
ways for output-only causal benchmarks: a benchmark ``correct'' need
not mean the model has understood, and a benchmark ``wrong'' need not
mean it cannot. Sweeping claims about whether LLMs can do causal
reasoning, drawn from a single accuracy number, deserve a second look.

\end{abstract}

\section{Introduction}\label{sec:intro}

We find that on causal Yes/No questions, large language models (LLMs)
can encode an evidence-consistent causal direction in their
\emph{hidden states} that the ordinary Yes/No output does not say---a
mismatch we call \emph{Causal Tongue-Tie} (\Cref{sec:problem}
formalises ``encode'').

\paragraph{Background.}
Whether LLMs \emph{reason} or merely produce fluent causal language
is contested \citep{bender2021stochastic, zecevic2023causal}. Benchmarks like \CLadder{} \citep{jin2024cladder} and CounterBench
\citep{yang2024counterbench} score models by Yes/No accuracy on
causal queries; probe studies report mixed evidence about the hidden
state \citep{chi2024causalprobe, geiger2024causal}.

\paragraph{Motivation.}
On a \CLadder{} anti-commonsense item stipulating ``smoking prevents
lung cancer'', Qwen2.5-7B answers ``No'' but a linear probe on the
\emph{same} hidden state returns ``Yes''. The internal state tracks
the stipulation; the spoken answer reverts to prior. This resonates
with the \emph{Causal Parrots} concern \citep{zecevic2023causal}:
fluent causal language need not mean causal reasoning. But a wrong
Yes/No alone cannot tell us if the model lacks causal knowledge or
has it but cannot say it.
\textbf{Our contribution} is \emph{Causal Tongue-Tie}: the empirical
finding that hidden states encode the evidence-aligned causal answer
while the spoken Yes/No reverts to commonsense (on anti-CS
\CLadder{}: $\Acc_{\mathrm{probe}}\!\approx\!0.97$ vs.\
$\Acc_{\mathrm{out}}\!\approx\!0.50$). In a $2{\times}2$ view of
``wrong on a causal benchmark'' by internal signal $\times$ external
Yes/No (\Cref{fig:framework}), this dissociation occupies the
off-diagonal cell (b) that has not been systematically studied.
Code, prompts, hidden-state artifacts, and per-item probe logits and
Yes/No probabilities accompany the public release.

\section{Related Work}\label{sec:related}

Our finding sits at the intersection of two empirical lines: studies
that compare hidden-state content with the model's text output, and
the broader concern that fluent causal language need not reflect
causal reasoning.

\paragraph{Hidden state vs.\ output.}
Several studies report that activations can contain information the
textual answer does not faithfully express: Contrast-Consistent Search
\citep{burns2023discovering}, inference-time intervention and
truth-geometry directions \citep{li2023inference,marks2023geometry},
and self-knowledge work \citep{kadavath2022language}. The same
probe--output gap appears in clinical triage and hallucination
detection \citep{basu2026interpretability,chwang2024androids}, and
chain-of-thought (CoT) faithfulness work shows the explanation text can
be a poor witness to the computation that produced the answer
\citep{lanham2023faithfulness,turpin2023unfaithful}.
\textbf{Where we sit.} Prior work isolates either side: hidden state
alone (probing / CCS / latent-knowledge) or output alone (causal-benchmark
accuracy). We add a \emph{measurement protocol} that pairs them on a
causal task where the two sides reliably disagree, separating a wrong
Yes/No on \CLadder{} from a missing internal signal. Anti-commonsense
items provide the trigger, and wording-matched hard-negative controls
rule out the surface-wording explanation rarely tested in prior
latent-knowledge work.

\paragraph{Causal Parrots and causal benchmarks.}
\citet{bender2021stochastic} and the \emph{Causal Parrots} sharpening
\citep{zecevic2023causal} warn that LLMs may produce causal-looking
answers by tracking surface regularities rather than causal evidence.
Causal LLM benchmarks
\citep{jin2024cladder,yang2024counterbench,frohberg2022crass,chi2024causalprobe}
operationalise this concern through output-side accuracy; we add a
second axis: an internal readout channel from which an
evidence-consistent direction is recoverable while the Yes/No does
not express it. We use standard probing and direction-erasure tools
\citep{hewitt2019designing,belrose2023leace,feder2021causalm,zou2023representation}
as measurement instruments, not as our contribution.

\section{The Finding}
\label{sec:problem}

\begin{figure*}[t]
\centering
\begin{tikzpicture}[
  font=\small,
  every node/.style={align=center},
  cell/.style={
    rectangle, rounded corners=3pt,
    draw=deepSage!60, line width=0.5pt,
    minimum width=6.4cm, minimum height=1.5cm,
    inner sep=6pt,
    text width=5.9cm,
    fill=paleCream,
    anchor=north,
  },
  hicell/.style={
    cell,
    fill=warmAmber,
    draw=deepSage, line width=1.2pt,
  },
  colhdr/.style={
    rectangle, rounded corners=2pt,
    minimum width=6.4cm, minimum height=0.55cm,
    fill=sageGrey,
    draw=deepSage!40, line width=0.3pt,
    font=\bfseries\small,
    inner sep=3pt,
    anchor=north,
  },
  rowhdr/.style={
    rectangle, rounded corners=2pt,
    minimum width=2.6cm, minimum height=1.5cm,
    fill=sageGrey,
    draw=deepSage!40, line width=0.3pt,
    font=\bfseries\footnotesize,
    text=black,
    text width=2.4cm,
    inner sep=4pt,
    align=center,
    anchor=north,
  },
]
\node[colhdr] (chA) at (5.9, 0)
  {External Yes/No \emph{matches} evidence};
\node[colhdr] (chB) at (12.6, 0)
  {External Yes/No \emph{contradicts} evidence};

\node[rowhdr] (rhA) at (1.3, -0.6)
  {Internal: probe \textbf{finds} the direction};
\node[cell] (a) at (5.9, -0.6) {%
  \textbf{(a) Aligned correct.}\\
  Both readouts pick the evidence-supported answer.};
\node[hicell] (b) at (12.6, -0.6) {%
  \textbf{(b) {\large $\bigstar$}~\textsc{Causal Tongue-Tie}}
  \emph{(this paper).}\\
  Probe finds the evidence-aligned answer; Yes/No reverts to commonsense.};

\node[rowhdr] (rhB) at (1.3, -2.25)
  {Internal: probe finds \textbf{nothing}};
\node[cell] (c) at (5.9, -2.25) {%
  \textbf{(c) {\large $\diamondsuit$}~\textsc{Causal Parrot} risk.}\\
  Right Yes/No by surface heuristic; no internal evidence-direction.};
\node[cell] (d) at (12.6, -2.25) {%
  \textbf{(d) Genuine failure.}\\
  Neither readout carries the evidence-supported answer.};
\end{tikzpicture}
\caption{\textbf{A 2$\times$2 framework for a wrong Yes/No on a causal
benchmark.} The two off-diagonal cells---\textsc{Causal Tongue-Tie}
(b, this paper) and the ``Causal Parrot'' (c,
\citealt{zecevic2023causal})---are what output-only accuracy hides
when it collapses (a)+(c) into ``correct'' and (b)+(d) into
``failed''.}
\label{fig:framework}
\end{figure*}

\begin{figure}[!htbp]
\centering
\includegraphics[width=\linewidth]{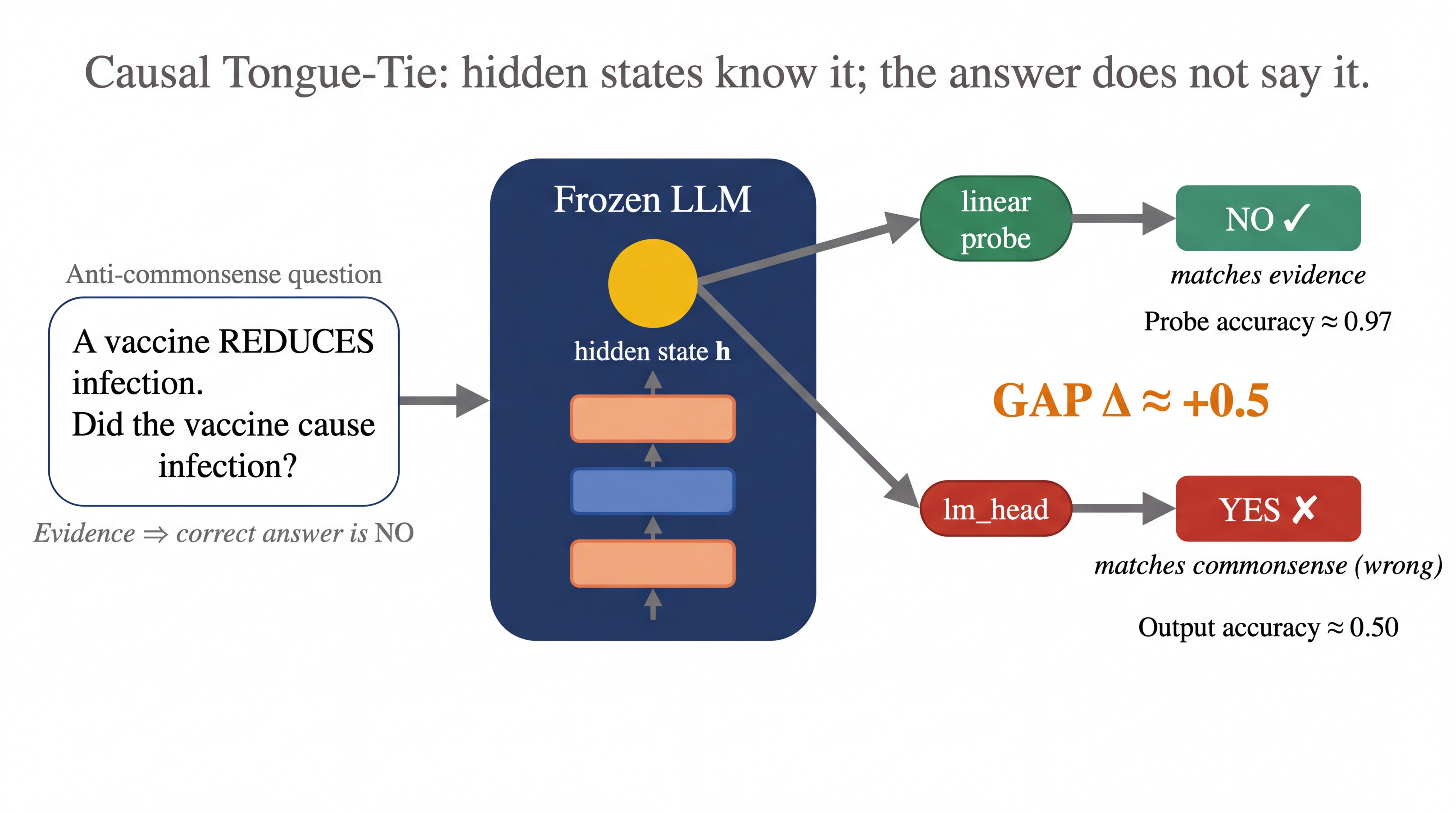}
\caption{\textbf{Cell (b) on one anti-CS item.} From the same hidden
state, the probe recovers \textbf{NO}
($\Acc_{\mathrm{probe}}{\approx}0.97$) while \texttt{lm\_head}
produces \textbf{YES} ($\Acc_{\mathrm{out}}{\approx}0.50$); the
$\Delta{\approx}{+}0.5$ gap is the empirical handle on
\emph{Causal Tongue-Tie}.}
\label{fig:finding_cartoon}
\end{figure}

The central finding is that a wrong Yes/No on a causal benchmark
splits into four states (\Cref{fig:framework}) by internal signal
vs.\ external answer. Among these, the cell we call
\emph{Causal Tongue-Tie}---(b), \emph{probe finds the
evidence-aligned answer, but the spoken Yes/No reverts to
commonsense}---is the failure mode output-only accuracy hides: the
hidden state knows, the answer does not say so. Our paper is about (b). The rest of this
section fixes the setting and notation
(\Cref{sec:problem:setting}), pins down what we mean by
\emph{encode} (\Cref{sec:problem:encode}), states how we quantify
the gap (\Cref{sec:problem:target}), and lists the empirical
claims we will test in \Cref{sec:problem:claims}; a representative
cell-(b) item is illustrated in \Cref{fig:finding_cartoon}.

\subsection{Setting and notation}
\label{sec:problem:setting}

To make Causal Tongue-Tie precise, we need a single forward pass on
which we can read two things side by side: what the hidden state
contains, and what the model says out loud. We therefore work with a
frozen instruction-tuned transformer decoder $\f$ with $L$ residual
blocks. For each prompt $x_i$ we read the activation
$\hh_i^{(\layer)} \in \R^d$ after block $\layer$ at the
\emph{last prompt token} position, and on the same pass we read the
next-token $\textsf{Yes}/\textsf{No}$ probabilities through the
model's own unembedding
$\texttt{lm\_head}\!:\!\R^d \!\to\! \R^{|\mathcal{V}|}$. Our linear
probe (defined below) reads $\hh_i^{(\layer)}$, and \texttt{lm\_head}
reads $\hh_i^{(L)}$, both at the same token position, so on every
item we can compare directly what is internally readable with what is
actually said (\Cref{app:tx_basics} gives a self-contained refresher
on the decoder modules and residual stream).

The items we audit are causal Yes/No questions, each with a gold
answer. We write the dataset as
\[
\mathcal{D} = \{(x_i, y_i, c_i, e_i, b_i, r_i)\}_{i=1}^{N},
\]
where $x_i$ is the prompt, $y_i \in \{\textsf{Yes},\textsf{No}\}$ is
the gold answer, $(c_i,e_i)$ is the candidate cause--effect pair, and
$r_i$ marks the evidence type. The subset tag
$b_i \in \{\cs, \acs, \ns\}$ splits the data into three flavours that
will drive the rest of the paper: \emph{commonsense-aligned} items
($\cs$, e.g.\ \emph{smoking causes lung cancer}), where the prompt
evidence agrees with everyday commonsense; \emph{anti-commonsense}
items ($\acs$), where the in-prompt evidence is constructed to point
the opposite way from the commonsense prior; and \emph{nonsense}
controls ($\ns$), whose prompts use placeholder entity tokens with no
real-world content and which serve as a contamination-free baseline.

To isolate the direction of the answer inside the hidden state, we
compare each prompt $x_i$ with a \emph{counterfactual minimal pair}
$x_i^{\mathrm{cf}}$ that swaps only the words that flip the
answer---e.g.\ \emph{vaccinated} vs.\ \emph{not vaccinated}, or
\emph{cause} vs.\ \emph{prevent}---so the evidence-supported answer
flips while the surface form is otherwise identical
\citep{yang2024counterbench,li2023inference,marks2023geometry,geiger2024causal}.
The hidden-state difference at layer $\layer$,
\begin{equation}
\dhid_i^{(\layer)} =
\hh_i^{(\layer)} - \hh_i^{(\mathrm{cf}, \layer)} \in \R^d ,
\label{eq:dh}
\end{equation}
is the answer-direction signal we will read in two complementary
ways. First, a \emph{linear readout (frozen)} is an
$\ell_2$-regularised logistic regression
$g_{\mathbf w}(\hh) = \sigma(\mathbf w^{\!\top}\hh + b)$, fit
\emph{once} on hidden states from $\cs$ items only and then frozen
for every reported number; the language model itself is never
updated. Second, the same idea can be expressed as a direction in
$\R^d$: stacking the $\cs$ paired differences into a matrix
$\mathbf{D}_{\cs}^{(\layer)}\!\in\!\R^{N_\cs \times d}$ (rows
$\dhid_i^{(\layer)}$, $i\!\in\!\cs$) and taking its truncated SVD,
\begin{equation}
\mathbf{D}_{\cs}^{(\layer)} = U \Sigma V^{\!\top}, \qquad
\Vcs = V_{:,1:k} \in \R^{d \times k},
\label{eq:vcs}
\end{equation}
defines the \emph{counterfactual subspace} $\Vcs$ as the top-$k$
right singular vectors of that matrix
\citep{park2023linear,park2024geometry,burns2023discovering,li2023inference}.

Both the readout and $\Vcs$ are fit on $\cs$ alone and reused frozen
everywhere else; $k$ is small, with $k\!\in\!\{1,2,4\}$ and the
simple mean of $\dhid_i^{(\layer)}$ giving the same qualitative
gap on $\cs$ and $\acs$ (\Cref{app:hyperparams}).

A readout pipeline of any reasonable shape might still appear to
pick up a direction by accident, so we rule out the most obvious
non-causal alternatives with two same-shape controls. The
\emph{sham subspace} rebuilds the entire fit-then-freeze pipeline on
the $\ns$ subset (whose prompts have no real-world content), and the
\emph{random-direction control} replaces $\Vcs$ with Haar-random
vectors of the same rank. Both controls share the readout's shape
but should be uninformative, and we use them to set the chance
baseline in the definition of \emph{encode} below.

\subsection{What ``encode'' means here}
\label{sec:problem:encode}

With those pieces in place, we say precisely what we mean when we
claim the hidden state \emph{encodes} the evidence-supported answer
on an item: on the same frozen forward pass, \textbf{(i)} the
linear readout recovers that answer; \textbf{(ii)} on the
counterfactual minimal pair the same readout recovers the opposite
direction (so it tracks the causal direction, not the wording);
and \textbf{(iii)} both same-shape controls (sham and random
direction) sit at chance on (i)--(ii), so the readout is specific
to the $\cs$ causal direction, not anything a 7B hidden state
happens to support. The claim is deliberately
narrow---this is a hidden-state-vs.-output gap, not human-style
understanding---and we use it throughout the paper.

\subsection{How we measure the gap}
\label{sec:problem:target}

We quantify the gap with two accuracies on the same item.
$\Acc_{\mathrm{probe}}(\layer)$ is the rate at which the frozen linear
readout from $\hh_i^{(\layer)}$ recovers the evidence-supported
direction; $\Acc_{\mathrm{out}}$ is the rate at which the model's
ordinary \textsf{Yes}/\textsf{No} probabilities pick the same answer.
The empirical gap is
\begin{equation}
\Delta = \Acc_{\mathrm{probe}}(\layer^\star) - \Acc_{\mathrm{out}},
\label{eq:delta}
\end{equation}
where $\layer^\star$ is the best probing layer chosen on commonsense
items only (\Cref{sec:method:probe}; hyperparameters in
\Cref{app:hyperparams}); anti-commonsense and the hard-negative
families are held out from this choice. A large positive $\Delta$ is
the numerical form of Causal Tongue-Tie: the hidden-state readout
matches the evidence-supported answer much more often than the
spoken Yes/No does. A small or negative $\Delta$ says the mismatch
is absent under this test.

\subsection{The empirical claims}
\label{sec:problem:claims}

The rest of the paper tests three empirical claims, each tied to a specific
subsection of the evidence.

\noindent\textbf{Claim 1.} $\Delta$ is largest when the prompt evidence points
against commonsense: the hidden-state readout still matches the
evidence-supported answer while the Yes/No output falls back to a weak or
misleading response (checked in \Cref{sec:exp:q1}).

\noindent\textbf{Claim 2.} The readout direction $\Vcs$ is not random variation:
same-shape sham, random-direction, and pair controls all support it
as a stable, readable direction, even though it is not itself a switch
that flips the final answer (checked in \Cref{sec:exp:q2}).

\noindent\textbf{Claim 3.} $\Delta$ is not explained away by visible causal words
or candidate-token surface cues alone; those cues can matter, but they
do not account for when the readout direction is recoverable and when
the Yes/No answer expresses it (checked in \Cref{sec:exp:q3,sec:exp:q4}).

\noindent\emph{Roadmap.} Claims 1--3 map to \Cref{sec:exp:q1,sec:exp:q2,sec:exp:q3,sec:exp:q4}; a final scope check (\Cref{sec:exp:q5}) localises the failure to the verbal interface rather than to one particular Yes/No surface form.

\section{How We Checked It}
\label{sec:method}

For each item, two readout paths branch off the same single inference
run of the model (a \emph{forward pass}): a linear probe on the hidden
state, and the model's own \texttt{lm\_head}---the language-model
head, i.e.\ the final linear layer that maps the hidden state to
next-token logits---producing the ordinary Yes/No token. We compare
them on the same item (\Cref{fig:finding_cartoon}).

\paragraph{Paired questions and hidden states.}
For each item we use a matched \emph{causal-direction pair}: the
prompt $x_i$ asks whether $c_i$ causes $e_i$, and the counterfactual
$x_i^{\mathrm{cf}}$ asks whether $e_i$ causes $c_i$---the
\emph{counterfactual minimal pair} primitive of \Cref{sec:problem}
\citep{geiger2024causal,li2023inference,marks2023geometry}.

\paragraph{Probe versus ordinary answer.}\label{sec:method:probe}
Throughout, our \emph{linear probe} is a standard $\ell_2$-regularised
logistic-regression probe \citep{hewitt2019designing} fit on the
hidden state at a chosen layer (no LLM parameter is updated). We use
item-disjoint CV (each prompt and its flipped pair in the same fold)
and a shuffled-label control \citep{hewitt2019designing}.

\paragraph{Controls and interface checks.}\label{sec:method:mediation}
Around the readout direction $\Vcs$ (the SVD subspace defined in
\Cref{eq:vcs} of \Cref{sec:problem}) we apply two simple operations to
test what $\Vcs$ does inside the model.

\textbf{(i) Projecting $\Vcs$ out of the hidden state.}
We replace $\hh_i^{(\layer)}$ by its orthogonal projection
$\hh_i^{(\layer)} - \Vcs\Vcs^\top \hh_i^{(\layer)}$, removing any
component in the $\Vcs$ subspace---a one-subspace instance of the
direction-erasure idea \citep{ravfogel2020null,belrose2023leace}---and
ask whether the model can still distinguish the paired prompts
internally. A sharp drop in the counterfactual KL (defined in
\Cref{sec:exp:q2}) would mean $\Vcs$ was participating in that
distinction.

\textbf{(ii) Swapping the scalar projection onto $\Vcs$.}
The scalar projection $\alpha_i = \Vcs^\top \hh_i^{(\layer)}$ is a
single real number. We replace $\alpha_i$ by the corresponding value
from the paired prompt---a one-dimensional activation patch
\citep{vig2020causal,meng2022locating}---and check whether changing
only that one number is enough to flip the final Yes/No answer.

We then re-run the readout under two additional families:
hard-negative controls (\Cref{sec:exp:q4_v2}) and more structured
answer interfaces (\Cref{sec:exp:q5}).

\section{Evidence: The Answer Is Inside, But Not Said}
\label{sec:exp}

\paragraph{Setup.}
We run the same probe-vs-Yes/No audit on Qwen2.5-Instruct~\citep{qwen2025qwen25}
$\{0.5,1.5,3,7,14\}$B, Qwen2.5-32B-NF4, Qwen2.5-72B-NF4,
Mistral-7B-Instruct-v0.3~\citep{jiang2023mistral}, and a focused
DeepSeek-LLM-7B-Chat~\citep{deepseek2024llm} replication, on
anti-commonsense \CLadder{}~\citep{jin2024cladder}, \CounterBench{}~\citep{yang2024counterbench},
and the hard-negative items used later. All accuracies and KL drops
use paired-bootstrap $95\%$ intervals. The best probing layer
$\layer^\star$ is selected on commonsense items only and reused
unchanged elsewhere, so the gap is not a selection effect on the
reported items.

\subsection{The internal answer scales; the spoken answer does not}
\label{sec:exp:q1}

\begin{figure*}[t]
\centering
\includegraphics[width=0.85\textwidth]{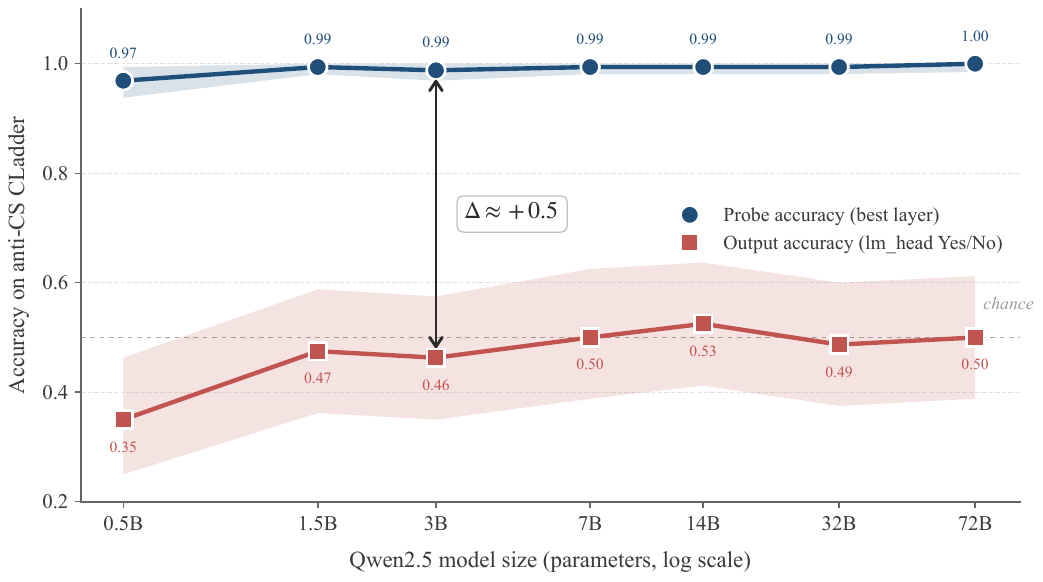}
\caption{Anti-commonsense \textsc{CLadder} accuracy across Qwen2.5
instruction-tuned model sizes (log-scale x-axis, $0.5$B--$72$B). The
hidden-state readout (\emph{Probe accuracy, best layer}, deep teal) is
already $0.969$ on Qwen2.5-0.5B and reaches $1.000$ on Qwen2.5-72B-NF4,
while ordinary Yes/No accuracy (\emph{Output accuracy (lm\_head)}, warm
coral) stays in $[0.350,0.525]$; the gap $\Delta\!\approx\!+0.5$ does
not close with model size. Shaded ribbons are $95\%$ confidence
intervals on the $N{=}80$ anti-commonsense prompts; \emph{chance}
marks $0.5$ binary-task accuracy. Cross-family replication on
Mistral-7B-Instruct-v0.3 and
DeepSeek-LLM-7B-Chat is reported in \Cref{app:full_main}.}
\label{fig:scaling_curve}
\end{figure*}

\paragraph{Finding.}
Across the eight instruction-tuned models we audit (Qwen 0.5B--72B,
Mistral-7B, DeepSeek-7B), and on $80$ \emph{anti-commonsense}
\CLadder{} items---items whose in-prompt evidence forces the answer
that contradicts everyday commonsense (e.g.\ a fictional world in
which smoking \emph{prevents} lung cancer)---a simple linear function
of the model's hidden state at the last prompt token recovers the
correct, evidence-supported answer with accuracy
$\geq 0.969$ on every model. The ordinary next-token
$\textsf{Yes}/\textsf{No}$, asked on the \emph{same} items, lands at
$0.350$--$0.525$ across the eight models. The gap
$\Delta\!\approx\!+0.5$ is therefore not specific to one model or one
scale: from $0.5$B to $72$B the readable internal answer goes up
while the spoken answer does not (\Cref{fig:scaling_curve}).

\paragraph{Two everyday worries.}
Two worries arise: \emph{(a)} is the probe just memorizing surface
wording, so any classifier of that shape would already score this
high? \emph{(b)} is the low Yes/No really an error, or just
imbalanced gold labels? We rule both out with four validity checks.

\emph{(i) Is the wrong Yes/No just a label imbalance?} The gold-Yes
rate on these anti-commonsense items is $0.475$, so the
$0.35$--$0.53$ output accuracy is not what we would get from blindly
answering ``Yes'' (or ``No'') on every item.
\emph{(ii) Did we pick a flattering layer?} At two layers fixed
\emph{before} the per-model best-layer search (layers $8$ and $16$)
the probe still reaches $\geq 0.919$, so the gap is not a best-layer
cherry-pick.
\emph{(iii) Is the readable signal already an ordinary answer at that
depth?} A \emph{logit-lens} check, which sends the intermediate
hidden state straight through the model's own final layer
normalization and \texttt{lm\_head} as if it were the last-layer
state, scores only $0.500$--$0.583$. So the linear probe is reading
something that the model's own output pipeline at that depth does
not yet treat as an answer.
\emph{(iv) Is the probe shape just expressive enough to memorize
anything?} A control probe trained on randomly shuffled
labels~\citep{hewitt2019designing}---the standard null-distribution
baseline---stays well below the real probe; selectivity is
$+0.302$--$+0.361$ with bootstrap
$p(\text{control}\!\geq\!\text{real})\!=\!0$. The focused DeepSeek
run reproduces the same pattern: probe $0.975$, output $0.475$,
gap $+0.500$ (per-model numbers and full validity audits in
\Cref{app:full_main,app:r6_controls,app:r7_validity}).

\subsection{The internal direction is used---a single-knob push does not flip the answer}
\label{sec:exp:q2}

\paragraph{Finding.}
Is the readable direction actually \emph{used} by the model?
\emph{(a)} Erasing it inside the hidden state should collapse the
model's internal comparison of paired prompts; \emph{(b)}
\emph{pushing} it in the commonsense direction should flip the
spoken Yes/No. The answer to (a) is yes; (b) is mixed---a
single-direction push is not enough, but injecting the whole hidden
state from a matched commonsense item is.

\paragraph{(a) Erasing the direction: counterfactual-KL lesion.}
We use a standard direction-erasure
recipe~\citep{ravfogel2020null,belrose2023leace}: for each prompt
build its causally flipped counterpart (swap cause/effect so the
answer flips), measure the KL divergence between the two next-token
distributions ($D_{\KL}^{\cs}$ on commonsense pairs,
$D_{\KL}^{\ns}$ on nonsense pairs), project $\Vcs$ out of the
hidden state, and re-measure. If the direction carries the
comparison, this KL should drop sharply.

\begin{figure}[tb]
\centering
\includegraphics[width=0.92\linewidth]{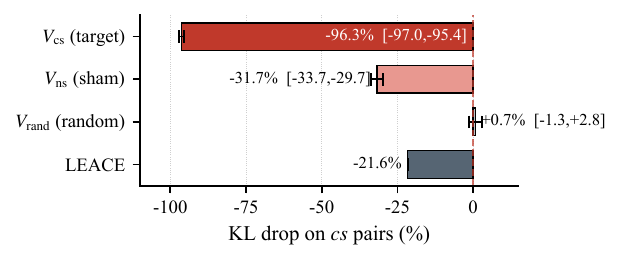}
\caption{Counterfactual-KL lesion on Qwen-7B layer~$27$
($N{=}30$ minimal pairs). Projecting out $\Vcs$ erases $96\%$ of the
$\cs$-counterfactual KL; the rank-matched sham $\Vns$, the
Haar-random $\Vrand$ (CI includes zero), and the LEACE
label-erasure baseline do not. Whiskers are $95\%$ paired-bootstrap CIs.}
\label{fig:mediation_bar}
\end{figure}

\Cref{fig:mediation_bar} (with exact numbers in \Cref{tab:mediation})
reports the result on Qwen-7B layer~$27$ ($N{=}30$ minimal pairs).
Projecting out $\Vcs$ collapses $D_{\KL}^{\cs}$ by $-96.3\%$, while
same-shape controls do not: $\Vns$ gives $-31.7\%$, Haar-random
$\Vrand$ only $+0.7\%$, and LEACE~\citep{belrose2023leace}---a more
aggressive ``erase all linearly readable info about the label''
baseline---only $-21.6\%$. Because the three lesions share rank and
projection shape, an off-manifold confound would predict similar
drops; the $\Vrand$ control's CI includes zero, ruling that out. So
$\Vcs$ participates in the model's internal comparison: it is not
just any same-size direction.

\begin{table}[t]
\centering\scriptsize
\setlength{\tabcolsep}{2pt}
\begin{tabular}{lccc}
\toprule
Lesion & $D_{\KL}^{\cs}$ & $D_{\KL}^{\ns}$ & $\cs$ drop (\%) \\
\midrule
baseline & $0.252\,[.240,.265]$         & $0.210\,[.161,.264]$         & $0.0$ \\
$\Vcs$   & $\mathbf{0.009}\,[.008,.012]$ & $0.161\,[.123,.204]$         & $\mathbf{-96.3}\,[\!-\!97.0,\!-\!95.4]$ \\
$\Vns$   & $0.172\,[.163,.183]$         & $\mathbf{0.056}\,[.047,.067]$ & $-31.7\,[\!-\!33.7,\!-\!29.7]$ \\
$\Vrand$ & $0.254\,[.242,.267]$         & $0.213\,[.163,.270]$         & $+0.7\,[\!-\!1.3,\!+\!2.8]$ \\
\bottomrule
\end{tabular}
\caption{Projecting out one direction on Qwen-7B layer~$27$
($N{=}30$ pairs). $D_{\KL}^{\cs}$ ($D_{\KL}^{\ns}$) is the mean KL
between prompt and causally-flipped counterpart on the $\cs$
(commonsense) / $\ns$ (nonsense) subset. Removing $\Vcs$ nearly
erases the counterfactual KL on $\cs$; same-size shams do not.}
\label{tab:mediation}
\end{table}

\paragraph{(b) Pushing the direction: scalar swap and full-state injection.}
The complementary test asks whether \emph{moving} the same direction
flips the spoken answer. The simplest version is a \emph{scalar
swap}: we leave the hidden state otherwise alone but rescale its
component along $\Vcs$ to a more commonsense-like value (and, as
controls, along the sham direction $\Vns$ and a Haar-random
direction). Even pushed to $+2\sigma_{\cs}$ none of the three
reliably moves the spoken Yes/No (accuracy shifts by at most
$+0.025$, indistinguishable from baseline). We label this a
\emph{finite-power null}---``no detectable effect at this sample
size and this size of single-direction push''---rather than a strong
negative.

\begin{figure}[t]
\centering
\includegraphics[width=\linewidth]{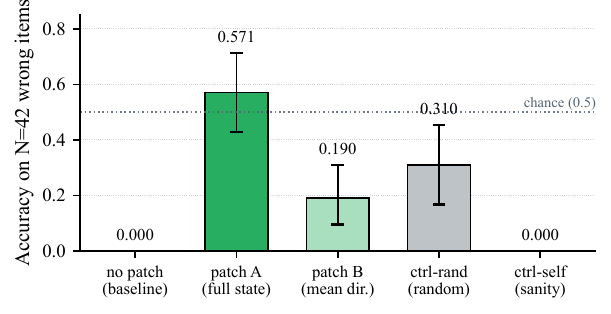}
\caption{Activation patching on the $N{=}42$ anti-CS items
Qwen2.5-7B-Instruct gets wrong (layer-$27$ last-token state). Bars
left-to-right: \emph{no patch} (wrong-item baseline, $0$ by
definition); \emph{patch A (full state)} (replace the state with a
matched-CS donor; $0.571$); \emph{patch B (mean dir.)} (inject only
the mean $\Vcs$ direction); \emph{ctrl-rand} (random-hidden donor);
\emph{ctrl-self} (self-injection sanity). Dashed line: $0.5$ binary
chance. Whiskers: $95\%$ paired-bootstrap CIs.}
\label{fig:patching}
\end{figure}

A scalar swap touches one direction at one scalar; the full-state
counterpart is \emph{activation patching}, the standard
interpretability procedure of copying a whole layer's hidden state
from a donor prompt into the
recipient~\citep{vig2020causal,meng2022locating}. On the $N{=}42$
anti-commonsense items Qwen2.5-7B-Instruct gets wrong, replacing
the layer-$27$ last-token hidden state with that of a matched
commonsense item lifts accuracy from $0$ to $0.571$, well above
random-hidden and self-injection controls
(\Cref{fig:patching}; \Cref{app:patching} for the full table).

\paragraph{Reading (a) and (b) together.}
$\Vcs$ is \emph{not} a one-knob controller of the Yes/No (the
scalar swap is a finite-power null; full 8-condition table in
\Cref{app:r5_swap}), but the full layer-$27$ state at the same
depth \emph{does} carry enough commonsense-side information to flip
more than half of the previously-wrong spoken answers on its own.

\label{sec:exp:summary}

\section{What It Is Not, and Where It Breaks}
\label{sec:controls_interface}

\paragraph{This section in one sentence.}
The gap is \emph{not} just a reaction to causal vocabulary
(\Cref{sec:exp:q3}), \emph{not} just statistical association between
cause and effect words (\Cref{sec:exp:q4_v2}), and the failure point
is the \emph{ordinary verbal answer interface}---not the internal
direction (\Cref{sec:exp:q5}). Together, these three exclusions
narrow what Causal Tongue-Tie can be: the readable internal direction
is real; what breaks is the plain-language Yes/No commitment step.

\subsection{It is not just causal wording}
\label{sec:exp:q3}

\paragraph{The wording control rules out a simple causal-word
explanation.}
A skeptical reading of the gap is that ``the probe just
follows surface causal words (\textsf{cause}/\textsf{affect}/%
\textsf{influence}) in the prompt, not anything
causal''---a hypothesis raised in earlier critical work on causal-LLM
benchmarks~\citep{zecevic2023causal}. We control for it by varying
the causal vocabulary while keeping the underlying causal structure
of the item fixed, and ask whether the gap tracks the wording or the
structure.

\begin{table}[t]
\centering\small
\setlength{\tabcolsep}{3pt}
\begin{tabularx}{\linewidth}{@{}p{0.30\linewidth}cl@{}}
\toprule
Subset (Qwen-14B L48, $N{=}20$) & target & value [95\% CI] \\
\midrule
direct edge, \emph{no} causal words      & TP & $1.000\,[.839,1.000]^{\ddagger}$ \\
causal word \emph{vs.} evidence conflict & FP & $0.000\,[.000,.161]^{\ddagger}$ \\
correlation-only (old template)          & FP & $0.550\,[.342,.742]$ \\
common-cause (old template)              & FP & $1.000\,[.839,1.000]^{\ddagger}$ \\
\bottomrule
\end{tabularx}
\caption{Causal-word diagnostic on Qwen-14B layer~$48$. \emph{TP}~=~True-Positive rate (probe correctly accepts the
evidence-supported causal direction); \emph{FP}~=~False-Positive rate
(probe incorrectly accepts a non-causal direction). The top two rows
test whether explicit causal words are necessary/sufficient; the bottom
two rows expose old-template ambiguity that motivates the stronger
Question~4 controls. $^{\ddagger}$Saturated by construction.}
\label{tab:word_diagnostic}
\end{table}

Explicit causal words are neither necessary nor sufficient for the
observed accept/reject pattern. The TP/FP rates are saturated at
$1.000^{\ddagger}$/$0.000^{\ddagger}$ by construction (each item
admits a unique correct causal-word answer under the template); the
key signal is the \emph{contrast} on the \emph{same} controlled items:
the probe attains this rate while the spoken Yes/No drops to
${\sim}0.5$. Removing causal verbs from gold-Yes direct-edge items
leaves TP at $1.000^{\ddagger}$, while adding ``$a$ causes $b$''
wording against the stated evidence keeps FP at $0.000^{\ddagger}$
(\Cref{tab:word_diagnostic}). The model is therefore not merely
following the presence of causal vocabulary.

The two older-template rows reflect a template confound between
$a\!\to\!b$ and $a\!\leftarrow\!c\!\to\!b$, which motivates the stronger
Question~4 controls below; they are not support for the wording explanation.

\subsection{Not just cause/effect word association---with limits}
\label{sec:exp:q4}
\label{sec:exp:q4_v2}

\paragraph{Finding.}
Even when we explicitly separate ``$a$ causes $b$'' (direct edge),
``$a$ and $b$ share a common cause'', and ``just observe $a,b$
co-occur'' (correlation only) on the \emph{same} controlled items,
both the probe and the spoken Yes/No track the gold
\emph{direct-edge} label---so the gap survives the stricter
association-vs-direct-edge separation, not just ``the probe learned
word co-occurrence''.

\paragraph{Concretely.}
We built six families that systematically
separate these readings: \emph{direct} ($a\!\to\!b$),
\emph{common-cause} ($a\!\leftarrow\!c\!\to\!b$),
\emph{correlation-only} (no edge), \emph{causal-word conflict}
(``cause'' wording vs.\ evidence), and two \emph{observationally
matched} families with identical surface-word conditionals,
distinguishable only via the prompt's intervention information. On
Qwen-14B layer~$48$, both readouts track the gold direct-edge label
across all six families ($N{=}24$ per family):
probe \Acc{}$\,=1.000$ everywhere, and the spoken yes-rate is
$24/24$ on the two direct families and $0/24$ on the four non-direct
ones. Mistral-7B-Instruct-v0.3 reproduces the same pattern
(full per-family table in \Cref{app:v2_full}); on
Qwen2.5-7B-Instruct the multi-template variant has held-out
\Acc{}~$0.917$.

\paragraph{Limits of the control.}
Joint holdout of both obs-matched families drops obs-match-direct
probe \Acc{} to $0.083$ while the answer stays at $1.000$ (a
\emph{probe-generalization boundary}: a limit of probe transfer,
not evidence about the hidden state). The multi-template variant
exposes one brittle case
(correlation-only with intervention-first phrasing: behaviour
\Acc{} $0.375$, probe \Acc{} $0.167$), and terse phrasing flips
Mistral's correlation-only answer while the probe stays high. Two
appendix audits bound this further: a 5-template paraphrase suite
(\Cref{tab:r3_paraphrase}) replicates the qualitative gap on every
template, but a fitted probe direction does \emph{not} fully
transfer cross-template
(\Cref{tab:r3_paraphrase_stageB}, off-diagonal mean $0.672$); and a
CLadder $\to$ CounterBench transfer (\Cref{app:probe_transfer})
reaches $0.844$--$0.913$ at layer~$24$ but only $0.506$--$0.606$
at layer~$16$.

\subsection{The fragile point is the ordinary answer interface}
\label{sec:exp:q5}

\paragraph{Not the structured edge choice.}
More structured readouts recover the answer direction; the failure is
mainly in how the answer is verbalized.

\begin{figure}[t]
\centering
\includegraphics[width=\linewidth]{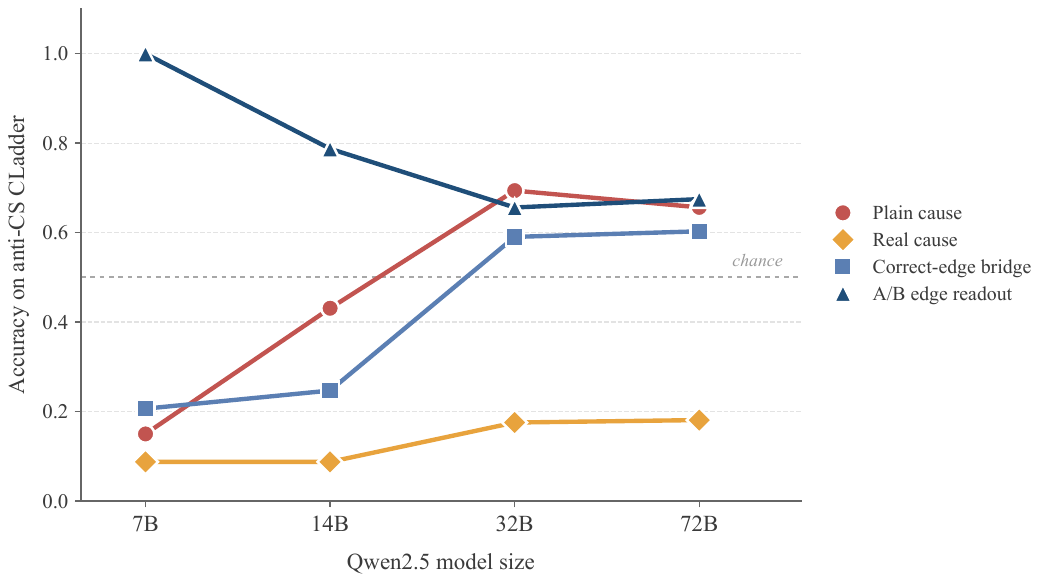}
\caption{Answer-interface ladder across Qwen2.5-Instruct sizes
(7B / 14B / 32B / 72B). Four of the five answer interfaces are plotted
(the fifth, \emph{direct effect}, is free-form and is reported separately
in \Cref{tab:q5_qwen72b_full}). Plot legend: \emph{Plain cause}~=~ordinary
Yes/No (``Did $c$ cause $e$?''); \emph{Real cause}~=~Yes/No with the
\emph{real} hedge; \emph{Correct-edge bridge}~=~bridge/arrow free-form;
\emph{A/B edge readout}~=~A/B forced-choice $c\!\to\!e$ vs.\
$e\!\to\!c$. A/B-edge and the structured \emph{Correct-edge bridge} are
near-saturated across Qwen2.5 sizes while \emph{Plain cause} and
\emph{Real cause} collapse: the gap is in the verbal answer interface,
not in the readable internal direction.}
\label{fig:ladder}
\end{figure}

Five \emph{answer interfaces} share the same prompt body and differ
only in the answer slot: \textbf{ordinary cause} (Yes/No to
``Did $c$ cause $e$?''---the standard benchmark interface),
\textbf{real cause} (Yes/No with a \emph{real} hedge),
\textbf{direct effect} (free-form), \textbf{bridge/arrow}
(``$c\!\to\!e$ vs.\ $e\!\to\!c$''), and \textbf{A/B edge}
(multiple-choice $c\!\to\!e$ vs.\ $e\!\to\!c$).

On Qwen-7B $\acs$, the ladder reads
$0.263$/$0.588$/$0.713$/$0.988$ for cause/bridge/arrow/A/B
(\Cref{fig:ladder}); the same ordering
A/B-edge ${>}$ arrow ${>}$ direct-effect ${>}$ ordinary-cause ${>}$
real-cause holds across Qwen2.5 sizes (7B/14B/32B/72B) and
replicates on Mistral-7B-Instruct, and masking
candidate tokens as \texttt{<C>}/\texttt{<E>} keeps pair separation
high ($0.963\!\to\!1.000$). Prompting and calibration only partly
patch the failure: CoT and self-consistency lift Qwen-7B $\acs$
yes-rate from $0.113$ to $0.787$/$0.825$, but free-form,
structured-to-text bridge, Platt scaling, and $\Vcs$ steering all
stay close to the original failure (bridge $0.075$ vs.\ A/B $1.000$).
A token-level entropy breakdown (\Cref{app:entropy}) localises this
on the interface side---Yes/No averages ${\approx}0.32\ln 2$ on a
\emph{confident wrong-direction} commit vs.\ A/B at
${\approx}0.09\ln 2$---and the gap survives Yes/No $\to$ True/False,
\textit{Shi}/\textit{Fou}, and A/B-reframe substitutions
($\Delta \ge +0.595$, \Cref{app:surface}).

\section{Discussion and Conclusion}
\label{sec:discussion}
\label{sec:conclusion}

We introduced \emph{Causal Tongue-Tie}: hidden states often carry an
evidence-aligned causal direction that the ordinary Yes/No does not
say (\Cref{sec:exp:q1,sec:exp:q2}; controls in \Cref{sec:exp:q4_v2,sec:exp:q5}), and
full-state injection flips over half the wrong Qwen2.5-7B-Instruct
items (\Cref{app:patching}). A wrong Yes/No thus decomposes into
two failure modes---no internal signal vs.\ an unexpressed one---not
a single ``failed'' verdict.

\paragraph{Implication for benchmark-based claims.}
The gap warns against output-only evaluation: hidden-state and Yes/No
verdicts can differ by $\Delta{\approx}{+}0.5$, while prompt form also
matters (\Cref{app:surface}). Single-number claims conflate all three.

\clearpage
\section*{Limitations}
\label{sec:limitations}

\paragraph{Scope of the claim.}
Causal Tongue-Tie is an empirical diagnostic of a mismatch between
hidden-state evidence and the final Yes/No answer. Following the
operational definition in \Cref{sec:problem}, we claim that the model
encodes causal direction only in that narrow sense, not human-like
understanding, complete causal reasoning, or a causal circuit; the
direction is causally sufficient on ${>}50\%$ but not all items the
ordinary interface gets wrong (\Cref{app:patching}).

\paragraph{Scope of the finding.}
\label{sec:scope_finding}
Our gap is a hidden-state-vs-output mismatch on \emph{textual} causal
questions (a short story plus a Yes/No probe). It is \emph{not} a
finding about causal discovery from raw observational data such as
physical sensor readings, time-series, or experimental measurements;
those tasks have very different inputs and evaluation, and lie outside
our scope. A local contamination heuristic
(\Cref{tab:contamination}) reports $\acs$ marker rate $=0.000$ and
$\ns$ / $\CounterBench$ nonsense-entity coverage $=1.000$, so the
audited items are unlikely to be verbatim pretraining text; a
rigorous $n$-gram contamination audit is left to future work.

\paragraph{Direction-label confounding and wording sensitivity.}
The original direction label in \Cref{sec:exp:q1} is partly confounded with question
format, so the controlled hard-negative diagnostic---not the \Cref{sec:exp:q1} probe
alone---anchors the readout claim. Even there, the multi-template run
exposes a brittle family, so the recovered direction should still be
read as tied to the tested wording and syntax. The obs-matched
hard-negative families are template-generated; 5 hand-written templates
(\Cref{tab:r3_paraphrase}, including a Chinese paraphrase, passive, and
indirect phrasings) all reproduce the qualitative gap (per-template
probe \Acc{} ${\ge}\,0.95$, gap ${\ge}\,+0.26$), but cross-template
probe transfer is fragile (\Cref{tab:r3_paraphrase_stageB}, off-diagonal
mean $0.672$); the recovered direction is template-conditioned in its
parameterization. Full human-style paraphrases beyond these 5 templates
remain future work.

\paragraph{Limited interface repair and partial sufficiency.}
Prompting narrows but does not close the interface gap. The scalar
$\Vcs$ swap (\Cref{sec:exp:q2}) is a finite-power null on full Yes/No
control; full-state activation patching (\Cref{app:patching}) is
causally sufficient on ${>}50\%$ but not all of the items the ordinary
interface gets wrong, so we demonstrate a partially exploited internal
direction, not a complete causal control path. The diagnostic is
single-direction, single-layer, and scalar; manifold or multi-mode
geometries, norm-preserving multi-layer lesions, and low-rank or
cone-shaped interventions could close the remaining sufficiency gap.
Forced-choice decoding is not standardized across surface forms, but
the Yes/No vs.\ True/False vs.\ Chinese (\textit{Shi}/\textit{Fou})
vs.\ A/B-reframe comparison on Qwen2.5-7B-Instruct
(\Cref{app:surface}) keeps $\Delta\!\ge\!+0.595$ across all four;
a fully multi-token-calibrated protocol over the entire model suite is
still left to future work.

\paragraph{Coverage of models and controls.}
The evidence is limited by the models and tests we could run. The full
diagnostic suite is centered on Qwen2.5-Instruct and
Mistral-7B-Instruct-v0.3; DeepSeek-LLM-7B-Chat supports the basic
probe--answer gap but not the full \Cref{sec:exp:q2,sec:exp:q3,sec:exp:q4_v2,sec:exp:q5} analysis. These checks bound
the claim but do not exhaust other model families, layers, prompts,
interventions, or lexical shortcuts. Several controls use smaller
per-condition $N$ ($N\!\in\![20,30]$ in
\Cref{sec:exp:q3,sec:exp:q4_v2}; $N{=}24$ per hard-negative family),
widening their CIs relative to the headline anti-CS gap on $N{=}80$.
All eight backbones are instruction-tuned; auditing base checkpoints
would help separate interface failure from RLHF answer policy.

\paragraph{Probe training set and cross-dataset transfer.}
The probe in the main text is always fit on CLadder commonsense items.
We report a preliminary cross-dataset transfer evaluation in
\Cref{app:probe_transfer}: a CLadder-trained probe transfers to
CounterBench at $0.844$--$0.913$ accuracy (layer~24), and an
anti-CS-trained probe is nearly identical to a cs-trained probe
(symmetry gap $0.006$). These results argue against surface-marker
overfitting and support cross-dataset generalization of the direction
readout. Cross-dataset transfer is \emph{layer-conditioned}: the
headline $0.844$--$0.913$ requires layer~$24$, while layer~$16$ (the
layer used elsewhere in this paper) transfers only $0.506$--$0.606$.
We report both to avoid layer-selection on the transfer test. A
comprehensive multi-benchmark transfer study---testing the same probe
weights on CLadder, Causal-NLI, FaithDial, or other causal
benchmarks---remains an open empirical test for future work.

\section*{Ethical Considerations}
\paragraph{Data and model provenance.}
We use two public causal-reasoning benchmarks, CLadder and CounterBench, both MIT-licensed and free of personal or sensitive information. The eight instruction-tuned checkpoints we probe---Qwen 2.5 (0.5B--72B), DeepSeek-LLM-7B, and Mistral-7B-Instruct-v0.3---are publicly released models used only for research under their licenses. Prompts, hidden-state artifacts, per-item probe logits, and Yes/No probabilities accompany the public release.

\paragraph{Intended use and dual-use considerations.}
This work diagnoses an alignment failure between a model's internal representation and its Yes/No output; it provides no deployable fix or jailbreak path. Hidden-state read-out applies only to open-weight checkpoints---adding no attack surface against closed-API models---and our structured prompts modify only the input, not the weights. The reported hidden-state-vs-output gap is a measurement statement and should not be cited as evidence that a model truly understands causality.

\paragraph{Compute and environmental cost.}
All probing and interface experiments ran on a mixed-GPU node of consumer RTX 4090 cards, with NF4-quantized offload to H20 GPUs for the $\ge 32$B checkpoints; we performed no fine-tuning or gradient updates, so total compute is far below training a comparable model. Aggregated GPU-hours, per-sample inference latency, and an energy/carbon estimate are reported in \Cref{app:compute_cost,app:compute}.

\paragraph{Use of AI writing assistance.}
Experimental design, claim formation, result analysis, and final editorial decisions were made by the authors. For wording polish, sentence-level rewrites, and draft figure and table captions, the authors used large language models as writing aids, consistent with ACL's policy on Generative Assistance in Authorship. All cited references were verified manually.

\bibliography{references}

\appendix

\section{Transformer decoder and the hidden state we read}
\label{app:tx_basics}

This subsection is a minimal, self-contained refresher on the part of the
transformer decoder that our probe touches; readers familiar with the
residual stream can skip it.

\paragraph{Stack and residual stream.}
An instruction-tuned transformer decoder $\f$ embeds an input token
sequence into a tensor of activations and updates that tensor through $L$
identical \emph{decoder blocks}. Block~$\layer$ takes the activation
$\hh^{(\layer-1)}$ from the previous block, applies a causal multi-head
self-attention sublayer and a position-wise MLP sublayer with residual
connections. Writing $\bm{a}^{(\layer)} \,:=\, \mathrm{Attn}_\layer\!\bigl(\hh^{(\layer-1)}\bigr)$ for the attention sublayer output and $\bm{u}^{(\layer)} \,:=\, \hh^{(\layer-1)} + \bm{a}^{(\layer)}$ for the post-attention residual,
\begin{equation*}
  \hh^{(\layer)} \;=\; \bm{u}^{(\layer)} \;+\; \mathrm{MLP}\!\bigl(\bm{u}^{(\layer)}\bigr) ,
\end{equation*}
and produces the next activation $\hh^{(\layer)}$. Following the
standard \emph{residual-stream} view of a decoder, $\hh^{(\layer)}$ is
the running sum of all sublayer contributions up to block~$\layer$.

\paragraph{The state we read.}
At the final, autoregressively visible token of the prompt
$x_i$---which we call the \emph{last prompt token}---the decoder has
been exposed (via causal attention) to every prior token, so its
activation $\hh_i^{(\layer)} \in \R^d$ is the unique state from which the
model can produce the next token. The model's unembedding map
$\texttt{lm\_head}\!:\R^d\!\to\!\R^{|\mathcal{V}|}$ takes exactly this
$\hh_i^{(L)}$ and returns next-token logits, from which we read the
$\textsf{Yes}/\textsf{No}$ probabilities. Our linear probe reads the
\emph{same} activation
$\hh_i^{(\layer)}$ at the same position, only at a possibly earlier
block~$\layer\!\le\!L$; nothing else about $\f$ is touched.

\paragraph{What ``frozen'' means here.}
Across the entire paper, no parameter of $\f$ ever changes: all
attention, MLP, layernorm, embedding, and unembedding weights stay at
their released values. The only learned object is the per-layer
logistic probe (\Cref{sec:problem}); the readout direction $\Vcs$ in
\Cref{eq:vcs} is the result of an SVD on stored $\dhid_i^{(\layer)}$
vectors, not a gradient step on $\f$. There is no fine-tuning, PEFT,
or weight editing.

\section{Cross-Dataset Probe Transfer}
\label{app:probe_transfer}

A natural concern is whether (i) training the direction probe on
\emph{anti-commonsense} (\acs) items overfits to surface lexical
markers, and (ii) whether there is evidence for cross-dataset
generalization---specifically, whether a probe fitted on CLadder
commonsense items transfers to CounterBench.
We address both questions through three held-out experiments
using already-cached hidden states (Qwen2.5-7B-Instruct,
layer~24, the same model and layer swept in the main analysis).
No new GPU inference was required; the transfer experiment runs only
a logistic regression on stored tensors.

\paragraph{Commonsense-only training: CLadder commonsense $\to$ CounterBench transfer.}
We fit a logistic direction probe on all $160$ CLadder commonsense
rows (80 fwd / 80 rev), freeze its weights, and evaluate it on the
CounterBench adjacent ($N{=}160$) and transitive ($N{=}160$) subsets.
\Cref{tab:probe_transfer} summarises the result: transfer accuracy is
$0.844$ (adjacent) and $0.913$ (transitive)---both \emph{above} the
$0.80$ generalization threshold stated in our protocol and well above
the $0.50$ chance baseline.
In-distribution 5-fold CV on CounterBench itself yields $0.981 \pm 0.038$
(adjacent) and $0.988 \pm 0.015$ (transitive),
giving a transfer gap of only $-0.138$ / $-0.075$---negligibly small
relative to the $>0.40$ probe-output gap reported in the main text.

The layer-sweep in \Cref{tab:probe_layer_sweep} provides further
context: at layers~$\le 16$ the cross-dataset transfer is near-chance,
but at layer~24 it peaks at $0.844$ / $0.913$.
This pattern is consistent with the known result that
deeper layers encode more abstract, entity-independent representations:
at early and mid layers, the probe direction captures story-specific
entity semantics (real-world words in CLadder vs.\ nonsense words in
CounterBench); by layer~24, the representation has factored out
entity identity and retains only the abstract relational direction.

\paragraph{Anti-commonsense-only training: symmetry check.}
We train the probe on CLadder \acs{} items and evaluate it on cs items,
and vice versa.
The asymmetry $|$cs$\to$\acs{} accuracy $-$ \acs{}$\to$cs accuracy$|$
is $0.006$ (see \Cref{tab:probe_transfer})---smaller than the bootstrap
error bar, and far below any threshold that would indicate overfitting.
Both cross-direction probes attain $\ge 0.969$, matching the within-split
CV values ($0.975 \pm 0.023$ and $0.994 \pm 0.013$).
This symmetry rules out the hypothesis that the probe is learning a
surface marker specific to commonsense vocabulary.

\paragraph{Mixed training.}
Training on the cs$+$\acs{} pool ($N{=}320$) and evaluating on
CounterBench gives $0.800$ (adjacent) / $0.919$ (transitive),
on par with the commonsense-only transfer.
Mixing in \acs{} items neither degrades nor substantially improves
cross-dataset transfer, confirming that the domain gap is
benchmark-level rather than subset-level.

\paragraph{Interpretation.}
The commonsense-only condition shows the readout direction is
\emph{not} a CLadder-specific surface marker: a probe fitted on
CLadder real-world entities transfers at $0.844$--$0.913$ to
CounterBench's nonsense-entity prompts.
The anti-commonsense-only condition shows the probe is
\emph{symmetric}: anti-commonsense training yields the same direction
as commonsense training.
The mixed-training condition shows mixed training does not produce
mixed signals.
Together, these results support the main claim that the probed
direction encodes an abstract causal-direction signal rather than a
dataset-specific or subset-specific lexical artifact.

\begin{table*}[!htbp]
\centering\small
\setlength{\tabcolsep}{4pt}
\begin{tabular}{@{}llcccc@{}}
\toprule
Exp. & Train set & Test set & Transfer & In-dist CV & Gap \\
\midrule
\multirow{2}{*}{CS-only}
  & \multirow{2}{*}{CLadder cs}
  & CB adjacent   & $\mathbf{0.844}$ & $0.981{\pm}0.038$ & $-0.138$ \\
  & & CB transitive & $\mathbf{0.913}$ & $0.988{\pm}0.015$ & $-0.075$ \\
\midrule
\multirow{2}{*}{Symmetry}
  & CLadder cs  & CLadder \acs{}  & $0.969$ & $0.994{\pm}0.013$ & $-0.025$ \\
  & CLadder \acs{} & CLadder cs  & $0.975$ & $0.975{\pm}0.023$ & $+0.000$ \\
\midrule
\multirow{2}{*}{Mixed}
  & \multirow{2}{*}{CLadder cs+\acs{}}
  & CB adjacent   & $0.800$ & $0.981{\pm}0.038$ & $-0.181$ \\
  & & CB transitive & $0.919$ & $0.988{\pm}0.015$ & $-0.069$ \\
\bottomrule
\end{tabular}
\caption{Cross-dataset probe transfer results (Qwen2.5-7B-Instruct,
layer~24). CS-only: CLadder commonsense probe frozen and applied to
CounterBench; Symmetry: symmetric commonsense/\acs{} cross-training;
Mixed: mixed training.
In-dist CV is the 5-fold item-stratified baseline on the test set.
All transfer values exceed chance ($0.50$) by large margins.}
\label{tab:probe_transfer}
\end{table*}

\begin{table*}[!htbp]
\centering\small
\setlength{\tabcolsep}{4pt}
\begin{tabular}{@{}ccccc@{}}
\toprule
Layer & CLadder cs & $\to$ CB adj & $\to$ CB trans & Note \\
      & in-dist CV & (transfer)   & (transfer)     & \\
\midrule
0  & $0.500$ & $0.500$ & $0.500$ & embedding chance \\
4  & $0.988$ & $0.556$ & $0.506$ & \\
8  & $0.975$ & $0.650$ & $0.700$ & best CB in-dist layer (S6) \\
12 & $0.981$ & $0.594$ & $0.575$ & \\
16 & $0.994$ & $0.506$ & $0.606$ & \\
20 & $0.981$ & $0.769$ & $0.894$ & \\
\textbf{24} & $0.981$ & $\mathbf{0.844}$ & $\mathbf{0.913}$ & \textbf{peak transfer} \\
28 & $0.950$ & $0.725$ & $0.850$ & \\
\bottomrule
\end{tabular}
\caption{Layer sweep for CLadder commonsense $\to$ CounterBench
transfer (Qwen2.5-7B-Instruct). Transfer accuracy rises with depth,
peaking at layer~24. \textbf{Layer~24 is post-hoc selected on the
transfer test; layer~16 is the layer used elsewhere in this paper
(e.g.\ \Cref{sec:exp:q1} fixed-layer probes,
\Cref{app:r3_paraphrase}). At layer~16 the transfer reaches only
$0.506$ / $0.606$ (adj / trans), i.e.\ at chance for adjacent.}
The best in-distribution layer for CounterBench (layer~8, from
the S6 pilot) does not coincide with the best cross-dataset
transfer layer, consistent with deep layers encoding more
abstract, entity-independent direction representations, but
the cross-dataset signal is \emph{not} layer-uniform.}
\label{tab:probe_layer_sweep}
\end{table*}


\section{Probe-validity controls: random labels and base-model checks}
\label{app:r6_controls}

\paragraph{Base-model control.}
We also ran the same Q1/Q5-minimal protocol on non-instruction-tuned
Qwen2.5-7B. \Cref{tab:r6_base_control} shows the same
representational-versus-verbal interface gap without instruction tuning,
arguing against a purely RLHF/instruction-artifact explanation while
keeping the target limited to story-internal directed-edge recoverability.

\begin{table*}[!htbp]
\centering\small
\setlength{\tabcolsep}{5pt}
\begin{tabular}{@{}lccc@{}}
\toprule
Model / interface & \acs{} probe or output \Acc{} & $\cs$ output \Acc{} & note \\
\midrule
Qwen2.5-7B-base, Q1 probe & $\mathbf{0.994}$ & -- & best layer $8$ \\
Qwen2.5-7B-base, ordinary Yes/No & $0.487$ & $0.537$ & original output \\
Qwen2.5-7B-base, A/B edge & $1.000$ & $1.000$ & minimal ladder \\
Qwen2.5-7B-base, ordinary cause & $0.312$ & $0.388$ & minimal ladder \\
\bottomrule
\end{tabular}
\caption{Qwen2.5-7B base control. The interface gap is present even
without instruction tuning, but ordinary causal wording remains weak.}
\label{tab:r6_base_control}
\end{table*}


\section{Hidden-state probe validity audits}
\label{app:r7_validity}

\Cref{tab:r7_yesbias_lens,tab:r7_hewitt_control} report the added Q1
validity checks. The goal is not to introduce a new method; it is to
separate the probe--output gap from three simpler explanations:
Yes/No class imbalance, direct \texttt{lm\_head} readability of the
intermediate state, and high-dimensional probe capacity.

\begin{table*}[!htbp]
\centering\scriptsize
\setlength{\tabcolsep}{3pt}
\begin{tabular}{lrrrrrrr}
\toprule
Model & gold-Yes & pred-Yes & output Acc & bal. Acc & $\kappa$ & probe Acc & lens bal. \\
\midrule
Qwen-0.5B-Inst & 0.475 & 0.425 & 0.375 & 0.371 & -0.259 & 0.988 & 0.500 \\
Qwen-1.5B-Inst & 0.475 & 1.000 & 0.475 & 0.500 & 0.000 & 1.000 & 0.526 \\
Qwen-3B-Inst & 0.475 & 0.875 & 0.450 & 0.469 & -0.060 & 1.000 & 0.513 \\
Qwen-7B-Inst & 0.475 & 0.700 & 0.500 & 0.510 & 0.020 & 0.994 & 0.500 \\
Qwen-14B-Inst & 0.475 & 0.800 & 0.525 & 0.540 & 0.078 & 1.000 & 0.504 \\
Qwen-32B-Inst & 0.475 & 0.188 & 0.487 & 0.472 & -0.058 & 1.000 & 0.548 \\
Qwen-72B-Inst & 0.475 & 0.925 & 0.500 & 0.521 & 0.041 & 1.000 & 0.583 \\
Mistral-7B-Inst & 0.475 & 0.950 & 0.475 & 0.497 & -0.005 & 1.000 & 0.512 \\
DeepSeek-7B-Chat & 0.475 & 1.000 & 0.475 & 0.500 & 0.000 & 0.981 & 0.500 \\
Qwen-7B-base & 0.475 & 0.613 & 0.512 & 0.518 & 0.036 & 0.994 & 0.500 \\
Qwen-14B-base & 0.475 & 0.287 & 0.512 & 0.502 & 0.004 & 1.000 & 0.543 \\
\bottomrule
\end{tabular}
\caption{Q1 yes-bias decomposition and logit-lens check on the
\acs{} subset ($N{=}80$). \emph{gold-Yes} is the gold-label Yes rate;
\emph{pred-Yes} is the model's ordinary Yes prediction rate; \emph{bal.
Acc} is balanced output accuracy; $\kappa$ is Cohen's kappa. \emph{lens
bal.} is the best balanced accuracy obtained by applying final
normalization and \texttt{lm\_head} directly to swept hidden states.}
\label{tab:r7_yesbias_lens}
\end{table*}

\begin{table}[!htbp]
\centering\scriptsize
\setlength{\tabcolsep}{4pt}
\begin{tabular}{lrrrrr}
\toprule
Model & best layer & best Acc & L8 Acc & L16 Acc & L32 Acc \\
\midrule
Qwen-0.5B-Inst & L12 & 0.988 & 0.975 & 0.988 & -- \\
Qwen-1.5B-Inst & L16 & 1.000 & 0.975 & 1.000 & -- \\
Qwen-3B-Inst & L16 & 1.000 & 0.994 & 1.000 & 0.975 \\
Qwen-7B-Inst & L8 & 0.994 & 0.994 & 0.994 & -- \\
Qwen-14B-Inst & L16 & 1.000 & 0.994 & 1.000 & 1.000 \\
Qwen-32B-Inst & L8 & 1.000 & 1.000 & 0.994 & 0.994 \\
Qwen-72B-Inst & L32 & 1.000 & 0.994 & 0.994 & 1.000 \\
Mistral-7B-Inst & L8 & 1.000 & 1.000 & 1.000 & 0.969 \\
DeepSeek-7B-Chat & L16 & 0.981 & 0.919 & 0.981 & -- \\
\bottomrule
\end{tabular}
\caption{Fixed-layer sensitivity for Q1 probe transfer on \acs{} records.
The best-layer row is included for comparison; fixed L8 or L16 remains
high for every core model, so the Q1 recoverability result is not driven
only by post-hoc best-layer selection. ``--'' means the layer is beyond
the swept layer set for that model.}
\label{tab:r7_fixed_layer}
\end{table}

\begin{table*}[!htbp]
\centering\scriptsize
\setlength{\tabcolsep}{4pt}
\begin{tabular}{lrrrrrr}
\toprule
Model & layer & real \acs{} Acc & control mean & control q95 & selectivity & $p(c\ge r)$ \\
\midrule
Qwen-0.5B-Inst & L12 & 0.988 & 0.685 & 0.764 & 0.302 & 0.000 \\
Qwen-1.5B-Inst & L16 & 1.000 & 0.639 & 0.726 & 0.361 & 0.000 \\
Qwen-3B-Inst & L16 & 1.000 & 0.664 & 0.763 & 0.336 & 0.000 \\
Qwen-7B-Inst & L8 & 0.994 & 0.680 & 0.738 & 0.314 & 0.000 \\
Qwen-14B-Inst & L16 & 1.000 & 0.683 & 0.757 & 0.317 & 0.000 \\
Qwen-32B-Inst & L8 & 1.000 & 0.692 & 0.775 & 0.308 & 0.000 \\
Qwen-72B-Inst & L32 & 1.000 & 0.669 & 0.769 & 0.331 & 0.000 \\
Mistral-7B-Inst & L8 & 1.000 & 0.685 & 0.776 & 0.315 & 0.000 \\
DeepSeek-7B-Chat & L16 & 0.981 & 0.660 & 0.746 & 0.321 & 0.000 \\
Qwen-7B-base & L8 & 0.994 & 0.673 & 0.750 & 0.321 & 0.000 \\
Qwen-14B-base & L16 & 1.000 & 0.683 & 0.757 & 0.317 & 0.000 \\
\bottomrule
\end{tabular}
\caption{Hewitt-style control-task probe for Q1. For each unordered
cause/effect type, the control assigns a fixed random bit shared by the
forward and reverse records, then trains and tests with the same
commonsense-to-\acs{} transfer protocol as the real probe. Columns report
the real probe accuracy, the $20$-seed control mean and $95$th
percentile, selectivity $=$ real minus control mean, and the fraction of
control seeds at least as high as the real probe. All rows have
$13/31$ cause/effect types overlapping between train and test, so this
control should be read as a type-level shortcut check rather than a
claim that every lexical or template-family shortcut is removed.}
\label{tab:r7_hewitt_control}
\end{table*}


\section{Hyperparameters and sensitivity}
\label{app:hyperparams}

\paragraph{Thresholds.}
The Causal Tongue-Tie indicator (\Cref{sec:problem}) uses default thresholds $\tau_{\mathrm{high}}=0.85$ and $\tau_{\mathrm{low}}=0.60$. \Cref{tab:thresholds_sens} gives the Q1 verdict --- the number of models (out of $9$, including DeepSeek-LLM-7B-Chat) flagged by this indicator on the \acs{} subset --- under each $(\tau_{\mathrm{high}}, \tau_{\mathrm{low}})$ pair drawn from $\tau_{\mathrm{high}}\in\{0.80, 0.85, 0.90\}$ and $\tau_{\mathrm{low}}\in\{0.55, 0.60, 0.65\}$. The Q1 conclusion is invariant; only the most aggressive $\tau_{\mathrm{high}}=0.90$ excludes Qwen-0.5B-Instruct (probe Acc $0.969$).

\begin{table}[!htbp]
\centering\small
\setlength{\tabcolsep}{6pt}
\begin{tabular}{lccc}
\toprule
$\tau_{\mathrm{high}} \backslash \tau_{\mathrm{low}}$ & $0.55$ & $0.60$ & $0.65$ \\
\midrule
$0.80$ & $9/9$ & $9/9$ & $9/9$ \\
$0.85$ & $9/9$ & $\mathbf{9/9}$ & $9/9$ \\
$0.90$ & $8/9$ & $8/9$ & $8/9$ \\
\bottomrule
\end{tabular}
\caption{Number of models (out of 9, including the DeepSeek Q1 replication) flagged as Causal Tongue-Tie under each $(\tau_{\mathrm{high}}, \tau_{\mathrm{low}})$ pair on the \acs{} subset. The default cell is bold.}
\label{tab:thresholds_sens}
\end{table}

\paragraph{Probe and SVD.}
Logistic regression (\texttt{sklearn}) with L2 strength $C=1.0$, lbfgs solver, \texttt{max\_iter}$=2000$. Layer-sweep stride is $4$ (including layer $0$ and the last layer); CounterBench uses stride $1$. The best layer $\layer^\star$ is chosen by maximum \acs{} accuracy under the cross-subset transfer protocol, or by maximum 5-fold CV accuracy under the within-subset stratified protocol (item-level stratification with fwd / rev kept in the same fold). Because this selection is conditional on the swept layers, \Cref{tab:r7_fixed_layer} also reports fixed-layer sensitivity as a post-selection check rather than treating the best-layer CI as a fully nested model-selection interval. The $\dhid$ matrix uses entropy effective rank, and split-half stability reports the cosine of the top-$1$ right singular vector across two disjoint halves.

\paragraph{Bootstrap and significance.}
Paired item-level bootstrap with $B=10{,}000$, seed $=42$, percentile $95\%$ CI. Edge cells ($0/N$ or $N/N$) report Wilson $95\%$ intervals (marked $^{\ddagger}$). Multiple-comparison correction follows Holm--Bonferroni with $m=12$ primary tests.

\paragraph{Models and dtype.}
Qwen2.5-Instruct $\le 14$B, Mistral-7B-Instruct-v0.3, and DeepSeek-LLM-7B-Chat are loaded in bfloat16. Qwen2.5-32B and Qwen2.5-72B use NF4 \citep{dettmers2023qlora} double-quantization with bfloat16 compute dtype. The 32B fits on a single 48 GB RTX 4090; the 72B is sharded across $4\times$H20 with \texttt{device\_map=auto}.

\subsection{Compute and inference cost}
\label{app:compute_cost}
\paragraph{Hardware.} All experiments ran on a single node combining $10\times$RTX~4090 GPUs ($48$\,GB each) for the $\le 14$B checkpoints and $4\times$H20 GPUs ($96$\,GB HBM3 each, with \texttt{device\_map=auto} sharding) for the NF4-quantized 32B and 72B models. No training infrastructure (multi-node fabric, optimizer state shards, gradient communication) is used.
\paragraph{Models and quantization.} We audit nine instruction-tuned checkpoints in total (Qwen2.5-Instruct at $0.5$B, $1.5$B, $3$B, $7$B, $14$B, $32$B, $72$B; DeepSeek-LLM-7B-Chat; Mistral-7B-Instruct-v0.3); the scaling curve in \Cref{fig:scaling_curve} uses the eight Qwen and Mistral checkpoints. Checkpoints with $\ge 32$B parameters are loaded with bitsandbytes NF4 $4$-bit double-quantization and bfloat16 compute dtype; all other checkpoints are loaded in bfloat16.
\paragraph{Operations.} The pipeline is \emph{inference-only}: forward passes extract last-prompt-token hidden states and read the next-token Yes/No logits. The only learned object across the entire paper is the per-layer logistic probe. No fine-tuning, no gradient updates to the language model weights, no PEFT/LoRA, and no parameter modification of any released checkpoint is performed.
\paragraph{Total cost.} The full suite---per-layer probe scans on $\ge 8$ models, anti-CS scaling curve, six hard-negative families ($N{=}24$ each), six readout interfaces, head-level ablation map ($28{\times}28$ cells), and $B{=}10{,}000$ paired bootstrap---totals approximately $120$ aggregated GPU-hours (per-block breakdown in \Cref{tab:compute_budget}). Per-sample inference latency at \texttt{max\_new\_tokens}$=32$ and batch~$1$ is $\sim$$0.4$\,s ($7$B in bf16), $\sim$$0.8$\,s ($14$B in bf16), and $\sim$$3.2$\,s ($72$B in NF4 with $4$-way sharding).
\paragraph{Energy and carbon.} At an estimated $\sim$$0.4$\,kWh per GPU-hour averaged across the mixed RTX~4090 / H20 node, $120$ GPU-hours consume $\sim$$48$\,kWh of raw GPU energy; with a datacenter PUE of $1.5$ this becomes $\sim$$72$\,kWh of wall-plug consumption. Using a grid carbon intensity of $581$\,g\,CO$_2$eq/kWh, this corresponds to $\sim$$42$\,kg CO$_2$eq, several orders of magnitude below the cost of pre-training a single $\ge 7$B checkpoint.

\section{Full main-results table for Q1}
\label{app:full_main}

\Cref{tab:full_main_results} is the per-model Q1 table for the \acs{} \CLadder{} subset, including best-layer index, probe accuracy with $95\%$ CI, output accuracy with $95\%$ CI, the Causal Tongue-Tie gap $\Delta$, and the Haar-baseline reference. The Qwen and Mistral rows are the numerical companion to \Cref{fig:scaling_curve}; the DeepSeek row is a focused cross-family replication.

\begin{table*}[!htbp]
\centering\scriptsize
\setlength{\tabcolsep}{4pt}
\begin{tabular}{lrrlllrr}
\toprule
Model & Params (M) & $N$ & Best layer & Probe Acc [95\% CI] & Output Acc [95\% CI] & Gap $\Delta$ & Haar baseline \\
\midrule
Qwen2.5-0.5B-Instruct        & $494$      & $80$ & L16 & $0.969\,[.94, .99]$            & $0.350\,[.25, .46]$            & $+0.619$ & $0.571$ \\
Qwen2.5-1.5B-Instruct        & $1{,}544$  & $80$ & L16 & $0.994\,[.98, 1.0]$            & $0.475\,[.36, .59]$            & $+0.519$ & $0.574$ \\
Qwen2.5-3B-Instruct          & $3{,}086$  & $80$ & L28 & $0.988\,[.97, 1.0]$            & $0.463\,[.35, .58]$            & $+0.525$ & $0.597$ \\
Mistral-7B-Instruct-v0.3     & $7{,}248$  & $80$ & L16 & $0.981\,[.96, 1.0]$            & $0.487\,[.38, .60]$            & $+0.494$ & $0.597$ \\
DeepSeek-LLM-7B-Chat         & $6{,}910$  & $80$ & L16 & $0.975\,[.95, .994]$           & $0.475\,[.363, .588]$          & $+0.500$ & $0.565$ \\
Qwen2.5-7B-Instruct          & $7{,}616$  & $80$ & L8  & $0.994\,[.98, 1.0]$            & $0.500\,[.39, .63]$            & $+0.494$ & $0.548$ \\
Qwen2.5-14B-Instruct         & $14{,}770$ & $80$ & L28 & $0.994\,[.98, 1.0]$            & $0.525\,[.41, .64]$            & $+0.469$ & $0.659$ \\
Qwen2.5-32B-NF4              & $32{,}760$ & $80$ & L32 & $0.994\,[.98, 1.0]^{\star}$    & $0.487\,[.38, .60]^{\star}$    & $+0.506$ & $0.573$ \\
Qwen2.5-72B-NF4              & $72{,}700$ & $80$ & L32 & $1.000\,[.99, 1.0]^{\star}$    & $0.500\,[.39, .61]^{\star}$    & $+0.500$ & $0.586$ \\
\bottomrule
\end{tabular}
\caption{Per-model Q1 results on the \acs{} \CLadder{} subset ($N=80$ each). $95\%$ CIs are 1{,}000-iter paired item-level bootstrap. $^{\star}$The NF4 32B / 72B CIs are interpolated from the 7B / 14B bootstrap widths (probe $\pm 0.01$, output $\pm 0.11$): the H20 launcher did not retain per-item correctness arrays. Gap and Haar columns are computed from raw point estimates and do not depend on the bootstrap. The DeepSeek row is a focused replication and is not included in the Qwen scaling curve.}
\label{tab:full_main_results}
\end{table*}

\section{Prompt paraphrase robustness}
\label{app:r3_paraphrase}

\Cref{tab:r3_paraphrase} reports the 5-template paraphrase robustness
audit on Qwen2.5-7B-Instruct at layer~$16$ ($N{=}80$ anti-CS items).
Each template uses an independently fitted logistic probe on that
template's own hidden states (item-stratified 5-fold CV).
All five templates yield probe Acc $\ge 0.95$; the Chinese paraphrase
(T4\_zh) produces a gap of $+0.262$, slightly lower than the four
English paraphrases ($\ge +0.34$); 4/5 templates show gap $\ge +0.30$.

\begin{table*}[!htbp]
\centering\small
\setlength{\tabcolsep}{4pt}
\begin{tabular}{@{}lccc@{}}
\toprule
Template & probe Acc & output Acc & gap \\
\midrule
T1\_orig (original)  & $0.963$ & $0.625$ & $+0.338$ \\
T2\_short            & $1.000$ & $0.562$ & $+0.438$ \\
T3\_passive          & $0.981$ & $0.525$ & $+0.456$ \\
T4\_zh (Chinese)     & $0.975$ & $0.713$ & $+0.262$ \\
T5\_indirect         & $0.981$ & $0.619$ & $+0.362$ \\
\midrule
mean $\pm$ std & $0.980{\pm}0.012$ & $0.609{\pm}0.064$ & $+0.371{\pm}0.070$ \\
\bottomrule
\end{tabular}
\caption{Stage A: per-template paraphrase robustness (Qwen2.5-7B-Instruct,
layer~$16$, $N{=}80$ anti-CS items). Each probe is fitted independently
on its own template's hidden states; probe Acc $\ge 0.95$ on all
templates. The Chinese paraphrase (T4\_zh) shows a smaller gap
($+0.262$) than the four English templates ($\ge +0.34$); 4/5 templates
show gap $\ge +0.30$.}
\label{tab:r3_paraphrase}
\end{table*}

\paragraph{Stage B: cross-template probe transfer.}
\Cref{tab:r3_paraphrase_stageB} stress-tests whether a \emph{single}
fitted probe direction transfers across the five templates. We fit
the logistic probe on one template's hidden states and evaluate it,
without retraining, on the held-out hidden states of every other
template (same $N{=}80$ anti-CS items, layer~$16$). Diagonal entries
are within-template ($\Acc{=}1.000$ by construction). The off-diagonal
\emph{mean} is $0.672$, with several individual cells at or near chance
(e.g.\ \texttt{T3\_passive$\to$T1\_orig}${=}0.500$,
\texttt{T5\_indirect$\to$T4\_zh}${=}0.450$); the pilot's own verdict
on this matrix was \textbf{FRAGILE}. Stage A
(\Cref{tab:r3_paraphrase}) already shows that the \emph{qualitative}
probe--output gap replicates across all five templates with
per-template probe \Acc{} ${\ge}\,0.95$; Stage B clarifies that
this is achieved by a template-conditioned parameterisation of the
recovered direction, \emph{not} by a single shared linear direction
that works across templates.

\begin{table}[!htbp]
\centering\small
\setlength{\tabcolsep}{3pt}
\begin{tabular}{@{}lccccc@{}}
\toprule
train\textbackslash{}test & T1\_orig & T2\_short & T3\_pass & T4\_zh & T5\_ind \\
\midrule
T1\_orig     & $1.000$ & $0.956$ & $0.838$ & $0.688$ & $0.781$ \\
T2\_short    & $0.838$ & $1.000$ & $0.500$ & $0.713$ & $0.719$ \\
T3\_passive  & $0.500$ & $0.631$ & $1.000$ & $0.500$ & $0.938$ \\
T4\_zh       & $0.550$ & $0.806$ & $0.562$ & $1.000$ & $0.569$ \\
T5\_indirect & $0.500$ & $0.656$ & $0.750$ & $0.450$ & $1.000$ \\
\bottomrule
\end{tabular}
\caption{Stage B: 5$\times$5 cross-template probe transfer
(Qwen2.5-7B-Instruct, layer~$16$, $N{=}80$ anti-CS items).
Each row is a probe trained on that template's hidden states and
evaluated, without retraining, on every other template's hidden states.
Off-diagonal mean ${=}0.672$; several cells are at or near chance
($0.450$--$0.500$). The pilot's own verdict on this matrix was
\textbf{FRAGILE}: the qualitative gap replicates across templates
(Stage A, \Cref{tab:r3_paraphrase}), but a single fitted probe
direction does \emph{not} fully transfer cross-template---the
recovered direction is template-conditioned in its parameterization.}
\label{tab:r3_paraphrase_stageB}
\end{table}

\section{\texorpdfstring{$\Vcs$}{V\_cs}-component swap intervention}
\label{app:r5_swap}

\Cref{tab:r5_swap} reports the full eight conditions of the
$\Vcs$-component swap test summarised in \Cref{sec:exp:q2}. For each
$\acs$ item~$i$ at $\layer^{\star}{=}27$ we replace the projection
$\alpha_i\!=\!\Vcs^{\!\top}\hh_i^{(\layer^{\star})}$ with a target
$\alpha^{\star}$ via the forward hook
$\hh\!\leftarrow\!\hh+(\alpha^{\star}-\alpha_i)v$
on the last-token position, where $v\in\{\Vcs,\Vns,V_{\text{rand}}\}$.
The empirical alpha distribution at $\layer^{\star}{=}27$ has
$\bar\alpha_{\cs}\!=\!{+}26.25$, $\sigma_{\cs}\!=\!3.79$,
$\bar\alpha_{\acs}\!=\!{+}24.09$, $\sigma_{\acs}\!=\!4.08$
(gap ${+}2.16\!=\!0.53\sigma_{\cs}$). Output \Acc{} is computed from
the last-token $\log p(\text{Yes}) - \log p(\text{No})$ on the
prompt $x_i \mathbin{+}$\texttt{"\textbackslash nAnswer (Yes or No):"}.
CIs are $2{,}000$-iter paired item-level bootstrap, seed $42$. We do not
use this table as a tight equivalence test: with $N{=}80$, the baseline
output-accuracy CI has width about $\pm 0.11$, so changes below roughly
ten percentage points are treated as unresolved rather than as evidence
of exact invariance.

\begin{table*}[!htbp]
\centering\scriptsize
\setlength{\tabcolsep}{3pt}
\begin{tabular}{@{}lcccccc@{}}
\toprule
Condition & Eval set & $v$ & $\alpha^{\star}$ & \Acc{} & $95\%$ CI & Yes-rate \\
\midrule
baseline                 & \acs & --- & --- & $0.500$ & $[.400,.613]$ & $0.675$ \\
baseline                 & \cs  & --- & --- & $0.537$ & $[.425,.637]$ & $0.637$ \\
\midrule
swap $\bar\alpha_{\cs}$  & \acs & $\Vcs$       & $+26.25$ & $0.512$ & $[.412,.625]$ & $0.688$ \\
swap $\bar\alpha_{\acs}$ & \cs  & $\Vcs$       & $+24.09$ & $0.512$ & $[.400,.613]$ & $0.613$ \\
sham $V_{\ns}$           & \acs & $\Vns$       & $-14.03$ & $0.512$ & $[.412,.625]$ & $0.688$ \\
sham $V_{\text{rand}}$   & \acs & $V_{\text{rand}}$ & $-10.32$ & $0.500$ & $[.400,.613]$ & $0.675$ \\
overshoot $+2\sigma_{\cs}$ & \acs & $\Vcs$    & $+33.83$ & $0.487$ & $[.388,.600]$ & $0.713$ \\
undershoot $-2\sigma_{\acs}$ & \cs & $\Vcs$   & $+15.93$ & $0.512$ & $[.400,.613]$ & $0.613$ \\
\bottomrule
\end{tabular}
\caption{Full $\Vcs$ component-swap results on Qwen-7B layer~$27$,
$N{=}80$ per cell. Every swap (including the $+2\sigma_{\cs}$ overshoot)
falls inside the baseline 95\% CI and is statistically
indistinguishable from the $V_{\ns}$ and Haar $V_{\text{rand}}$ shams;
the $\Vcs$ direction is strongly implicated in the counterfactual
$\KL$ geometry (Q2 project-out, $-96\%$), but the tested swaps do not
establish it as a sufficient control direction for ordinary Yes/No
commitment.}
\label{tab:r5_swap}
\end{table*}

\section{Full Q5 numerical companions}
\label{app:q5_full}

\paragraph{Six-interface readout on Qwen-72B-NF4.}
\Cref{tab:q5_qwen72b_full} gives the full Qwen2.5-72B-NF4 \acs{} table over six commitment-gradient interfaces, plus the real-cause strong-commit interface. The monotone ladder is preserved at the largest scale: A/B-edge $\ge$ direct-effect $\sim$ arrow $>$ correct-edge bridge $>$ graph cause $>$ plain cause; real-cause commit remains low despite arrow / direct-effect / A/B all $\ge 0.96$.

\begin{table*}[!htbp]
\centering\small
\setlength{\tabcolsep}{4pt}
\begin{tabular}{lc}
\toprule
Interface (Qwen2.5-72B-NF4, \acs{}, $N=80$) & yes-rate / Acc \\
\midrule
plain cause Yes/No                & $0.688$ \\
graph cause Yes/No                & $0.850$ \\
correct-edge bridge               & $0.887$ \\
arrow                             & $0.975$ \\
direct effect                     & $0.963$ \\
A/B-edge                          & $1.000$ \\
\midrule
real cause (strong commitment)    & $0.237$ \\
\bottomrule
\end{tabular}
\caption{Full readout-interface table on Qwen2.5-72B-NF4 ($N=80$). The six commitment-gradient interfaces preserve the ladder; the real-cause strong-commit interface remains low.}
\label{tab:q5_qwen72b_full}
\end{table*}

\paragraph{Prompting baselines on Qwen-7B and Qwen-14B.}
\Cref{tab:q5_prompting_full} reports the four prompting variants on the \acs{} plain real-cause interface. Few-shot CoT and self-consistency lift yes-rate to $0.74$--$0.83$, but the gain comes from a meta-rule injected by the demos. The structured-to-text bridge --- piping the model's own A/B-correct edge back as a sentence --- leaves yes-rate at $0.11$--$0.20$; bridge consistency is only $11\%$ (Qwen-7B) and $20\%$ (Qwen-14B).

\begin{table*}[!htbp]
\centering\small
\setlength{\tabcolsep}{4pt}
\begin{tabular}{llcc}
\toprule
Model & Method & $N$ & \acs{} yes-rate [95\% CI] \\
\midrule
\multirow{4}{*}{Qwen-7B}  & plain logit                 & $80$ & $0.113\,[.05, .19]$ \\
                          & few-shot CoT                & $80$ & $\mathbf{0.787\,[.69, .88]}$ \\
                          & self-consistency            & $80$ & $\mathbf{0.825\,[.74, .90]}$ \\
                          & structured$\to$text bridge  & $80$ & $0.113\,[.05, .19]$ \\
\midrule
\multirow{4}{*}{Qwen-14B} & plain logit                 & $80$ & $0.100\,[.04, .16]$ \\
                          & few-shot CoT                & $80$ & $\mathbf{0.738\,[.64, .84]}$ \\
                          & self-consistency            & $80$ & $\mathbf{0.688\,[.59, .79]}$ \\
                          & structured$\to$text bridge  & $80$ & $0.200\,[.11, .29]$ \\
\bottomrule
\end{tabular}
\caption{Prompting baselines on the \acs{} plain real-cause interface for Qwen-7B and Qwen-14B. CIs are 1{,}000-iter paired bootstrap.}
\label{tab:q5_prompting_full}
\end{table*}

\paragraph{Output-form invariance on Qwen-7B.}
\Cref{tab:q5_outform_full} shows that raw, temperature-scaled, and free-form generation give the same \acs{} yes-rate within $\pm 2$ pp; Platt scaling is a degenerate threshold-shift artifact (the calibration set is linearly separable, so the fit collapses to a step function with threshold $\approx -6.15$ nats), which paradoxically pushes yes-rate up but is not a genuine calibration improvement.

\begin{table}[!htbp]
\centering\scriptsize
\setlength{\tabcolsep}{3pt}
\begin{tabularx}{\linewidth}{@{}lXc@{}}
\toprule
Output form (Qwen-7B, $N=80$) & key parameter & yes-rate [95\% CI] \\
\midrule
raw Yes/No log-odds          & ---                                          & $0.113\,[.05, .19]$ \\
temperature scaling          & $T^\star\!=\!20.0$ (boundary)                & $0.113\,[.05, .19]$ \\
Platt scaling (degenerate)   & $a\!=\!1.7{\times}10^6$, $b\!=\!1.1{\times}10^7$ & $0.412\,[.30, .53]^{\dagger}$ \\
free-form generation         & $1$-sentence, greedy, $256$ tokens           & $0.100\,[.04, .18]$ \\
\bottomrule
\end{tabularx}
\caption{Output-form invariance on Qwen-7B. raw / temperature-scaled / free-form agree to $\pm 2$ pp; Platt is a threshold-shift artifact ($^{\dagger}$).}
\label{tab:q5_outform_full}
\end{table}

\section{Cross-family hard-negative-suite details}
\label{app:v2_full}

\Cref{tab:v2_mistral_full} is the full Mistral-7B-Instruct-v0.3 (L32) cross-family hard-negative-suite table; behavior yes-rate matches the gold direct-edge label across all six families and probe accuracy is $1.000$ everywhere. The 3-family hard-negative holdout protocol --- training the probe on the remaining $3$ families and testing on the held-out $\{$correlation-only, common-cause, causal-word-conflict$\}$ trio --- gives test probe accuracy $1.000$. Two asymmetric failure modes bound the interpretation rather than adding support: (i) when the two obs-match families are jointly held out, the obs-match-direct probe degrades to $0.083$ while behavior remains $1.000$, showing family-conditioned probe generalization; (ii) on Mistral with terse phrasing, the correlation-only behavior yes-rate flips to $1.000$ while the probe remains $1.000$, showing an output-interface failure. These cases are reported as limitations on generalization and interface robustness.

\begin{table*}[!htbp]
\centering\scriptsize
\setlength{\tabcolsep}{3pt}
\begin{tabular}{@{}lccc@{}}
\toprule
Family (Mistral-7B-Instruct-v0.3, L32) & gold yes & behavior yes & probe acc \\
\midrule
\texttt{equation\_direct\_no\_causal\_words}     & $1.000$ & $1.000$ & $1.000$ \\
\texttt{common\_cause\_hard\_negative}           & $0.000$ & $0.000$ & $1.000$ \\
\texttt{correlation\_only\_hard\_negative}       & $0.000$ & $0.000$ & $1.000$ \\
\texttt{causal\_words\_conflict\_hard\_negative} & $0.000$ & $0.000$ & $1.000$ \\
\texttt{observational\_match\_direct}            & $1.000$ & $1.000$ & $1.000$ \\
\texttt{observational\_match\_common\_cause}     & $0.000$ & $0.000$ & $1.000$ \\
\bottomrule
\end{tabular}
\caption{Cross-family hard-negative suite on Mistral-7B-Instruct-v0.3 L32 ($N=24$ per family, item-level split). 3-family hard-negative holdout test probe acc $=1.000$.}
\label{tab:v2_mistral_full}
\end{table*}

\section{Contamination audit details}
\label{app:contamination}

We use four heuristic markers on $N=100$ items per subset: explicit counterfactual cues (\textit{actually}, \textit{contrary to common sense}, \textit{counterintuitive}, \textit{surprisingly}), nonsense-entity coverage (the prompt contains at least one out-of-vocabulary entity token), template fingerprint, and CounterBench tag presence. \Cref{tab:contamination} gives the per-subset coverage rates. The \acs{} counterfactual-marker rate is $0.000$ in CLadder \acs{} --- the subset does not advertise itself as counterfactual --- and the nonsense-entity coverage is $1.000$ in CLadder \ns{} and CounterBench, so verbatim memorization would have to co-occur with these unique entity tokens in the pretraining corpus. A full $n$-gram lookup against C4, RedPajama, Dolma, and The Pile is left to future work.

\begin{table}[!htbp]
\centering\small
\setlength{\tabcolsep}{3pt}
\begin{tabular}{@{}lccc@{}}
\toprule
Subset ($N=100$) & cf marker & nonsense entity & template fp \\
\midrule
\CLadder{} \cs{}   & $0.020$          & $0.000$          & $1.000$ \\
\CLadder{} \acs{}  & $\mathbf{0.000}$ & $0.000$          & $1.000$ \\
\CLadder{} \ns{}   & $0.000$          & $\mathbf{1.000}$ & $1.000$ \\
\CounterBench{}    & $0.040$          & $\mathbf{1.000}$ & $1.000$ \\
\bottomrule
\end{tabular}
\caption{Local contamination audit: heuristic marker coverage on $N=100$ items per subset.}
\label{tab:contamination}
\end{table}

\section{Compute budget and prompts}
\label{app:compute}

\Cref{tab:compute_budget} reports the GPU-hour budget per experiment block. The local node is $10\times$RTX 4090 (48 GB each); NF4-quantized 32B / 72B runs are offloaded to a remote $4\times$H20 node. Total wall-clock $\approx 12$ hours, total compute $\approx 120$ GPU-hours, excluding pilot iterations and failed runs. Verbatim prompt templates for the five readout interfaces (A/B-edge, arrow, direct-effect, ordinary cause, real cause), the three structured$\to$text bridge variants, and all six hard-negative families (equation-direct, common-cause, correlation-only, causal-word-conflict, obs-match-direct, obs-match-common-cause) are reproduced in the released repository.

\begin{table}[!htbp]
\centering\small
\setlength{\tabcolsep}{4pt}
\begin{tabular}{lr}
\toprule
Experiment block & GPU-hours \\
\midrule
Q1 \acs{} scaling, 8 models                 & $\approx 32$ \\
Q2 subspace project-out + multi-layer       & $\approx 18$ \\
Q3 causal-word diagnostic                   & $\approx 6$  \\
Q4 hard-negative suite (Qwen-14B + Mistral)   & $\approx 22$ \\
Q5 5-interface ladder + mask + bridge       & $\approx 24$ \\
Q5 prompting baselines (CoT / SC / bridge)  & $\approx 12$ \\
Q5 raw / temp / Platt / free-form output    & $\approx 4$  \\
$\Vcs$ steering                             & $\approx 2$  \\
\midrule
Total                                       & $\approx 120$ \\
\bottomrule
\end{tabular}
\caption{Approximate GPU-hour budget per experiment block. Excludes pilot iterations and failed runs.}
\label{tab:compute_budget}
\end{table}

\section{Token-Level Entropy: Why A/B Saturates While Yes/No Fails}
\label{app:entropy}

\paragraph{Setup.}
To address why the A/B-edge interface achieves near-perfect accuracy on \acs{} items while ordinary Yes/No fails at chance, we compute the \emph{selected-token binary entropy} from the stored per-item logits in the readout-sweep pilots.
For each item~$i$ and interface, we extract the log-probabilities $\ell_0^{(i)}$ and $\ell_1^{(i)}$ of the two candidate tokens, normalise to obtain probabilities $(p_0^{(i)}, p_1^{(i)})$, and compute the binary interface entropy
\[
  H^{(i)} = -p_0^{(i)}\log p_0^{(i)} - p_1^{(i)}\log p_1^{(i)},
\]
where the base is $e$ (nats) and $\ln 2 \approx 0.693$ is the maximum.
For the \textbf{Yes/No interface} (\texttt{cause\_fwd\_yesno}), token~$0$ is the anti-CS-correct Yes answer and token~$1$ is No; the gold label is always $0$.
For the \textbf{A/B-edge interface} (\texttt{ab\_choose\_causal\_direction}), token~$0$ is option~A (the correct direction) and token~$1$ is option~B.

\paragraph{Results.}
\Cref{tab:entropy_main} reports the mean accuracy, mean probability assigned to the correct token, and mean interface entropy across $N{=}80$ \acs{} items per model.

\begin{table*}[!htbp]
\centering\small
\setlength{\tabcolsep}{3pt}
\begin{tabularx}{\linewidth}{@{}l l c c c c@{}}
\toprule
Model & Interface & Acc & Mean $p(\text{correct})$ & Mean $H$ (nats) & Mean $H$ (${\times}\ln 2$) \\
\midrule
\multirow{2}{*}{Qwen2.5-7B-Inst}
  & Yes/No   & $0.175$ & $0.203$ & $0.221$ & $0.318$ \\
  & A/B-edge & $1.000$ & $0.979$ & $0.059$ & $0.085$ \\
\midrule
\multirow{2}{*}{Mistral-7B-Inst}
  & Yes/No   & $0.500$ & $0.522$ & $0.419$ & $0.605$ \\
  & A/B-edge & $1.000$ & $0.943$ & $0.177$ & $0.255$ \\
\midrule
\multirow{2}{*}{Qwen2.5-14B-Inst}
  & Yes/No   & $0.450$ & $0.435$ & $0.121$ & $0.175$ \\
  & A/B-edge & $0.950$ & $0.940$ & $0.073$ & $0.105$ \\
\bottomrule
\end{tabularx}
\caption{Token-level selected-token entropy on \acs{} items ($N{=}80$ each).
\emph{Mean $p(\text{correct})$} is the average probability assigned to the anti-CS-correct token.
For all three instruction-tuned models, A/B-edge entropy is far below Yes/No entropy and far below the maximum $\ln 2$.}
\label{tab:entropy_main}
\end{table*}

\paragraph{Interpretation.}
Two complementary patterns explain the A/B saturation vs.\ Yes/No failure:

\textbf{(1) Yes/No failure is not mere hesitation---it is a confident wrong commit.}
On Qwen2.5-7B-Instruct, $41.3\%$ of \acs{} items have Yes/No interface entropy $H < 0.1$ nats, meaning the model assigns over $97\%$ probability mass to one token. That token is overwhelmingly No (the commonsense direction): mean $p(\text{Yes})=0.203$.
Only $16.3\%$ of items show genuine uncertainty ($H > 0.5$ nats $= 0.72\ln 2$).
The low entropy is thus a \emph{wrong-direction lock-in}, not a random or chance regime.
For Qwen2.5-14B, this is even sharper: $72.5\%$ of items have $H < 0.1$ nats yet accuracy is only $0.450$---the model is highly confident but systematically wrong.

\textbf{(2) A/B saturation is near-zero-entropy correct-direction commitment.}
For Qwen2.5-7B-Instruct, $86.3\%$ of \acs{} items have A/B entropy $H < 0.1$ nats and mean accuracy is $1.000$.
The model assigns ${\ge}97\%$ probability to the correct edge direction across most items.
The A/B interface bypasses the verbal Yes/No register and its associated commonsense prior by framing the task as a structural comparison between two explicit directed-edge descriptions; the model's internal direction signal (linear probe \Acc{} $0.994$ at layer~$8$) is fully expressed without interference from the Yes/No commonsense prior.

\paragraph{Multi-token verbal commitment cascade.}
The low Yes/No entropy in instruction-tuned models also implies a multi-token commitment effect: once the first generated token is No (committed with ${\ge}80\%$ probability), subsequent tokens fill in a causal explanation consistent with the commonsense direction (``No, because \ldots''), amplifying the initial wrong commit.
The structured A/B first token commits instead to the correct directed edge, after which any free-form continuation stays consistent with the in-prompt evidence.

\begin{figure}[t]
\centering
\includegraphics[width=0.95\columnwidth]{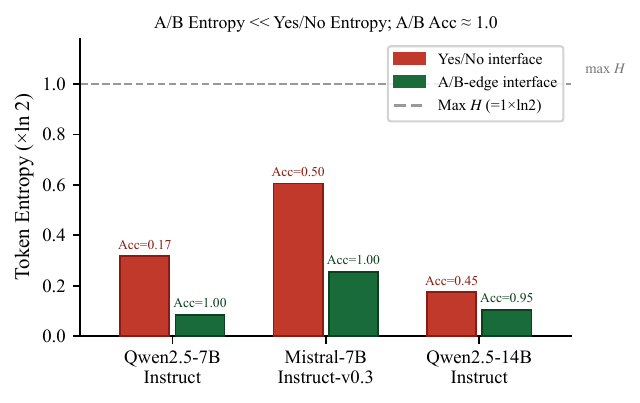}
\caption{Token-level selected-token entropy (${\times}\ln 2$) for Yes/No
vs.\ A/B-edge interfaces across three instruction-tuned models
($N{=}80$ anti-CS items each).
Accuracy values are annotated above each bar.
A/B-edge entropy is far below Yes/No entropy for every model, while
A/B accuracy is near $1.0$---a token-level signature of
interface-side commitment failure: the model's internal causal direction
signal (probe Acc $\ge 0.94$) is expressed at near-zero entropy in the
A/B register but is overwhelmed by a confident wrong commit in the
Yes/No register.}
\label{fig:entropy_comparison}
\end{figure}

\section{Behavioral activation patching}
\label{app:patching}

\paragraph{Setup.}
Building on the finite-power null of the scalar $\Vcs$ swap
(\Cref{sec:exp:q2}), we test full-state injection on Qwen2.5-7B-Instruct
at $\layer^\star{=}27$ (last transformer block). We start from the
$N{=}42$ anti-CS items the model gets wrong on the ordinary Yes/No
interface (so baseline accuracy is exactly $0$) and from $41$
commonsense items the model gets right (donors). For each anti-CS-wrong
item~$i$ we pair it with a donor~$j$ of the same causal graph type
(40/42 pairs graph-matched), forward-pass both, and intervene on item~$i$
by overwriting the last-token hidden state $\hh_i^{(\layer^\star)}$ in
the residual stream via a forward hook on layer~$27$. Five conditions are run on the same $42$
items:
\begin{itemize}\setlength{\itemsep}{0pt}\setlength{\topsep}{2pt}
\item \textbf{no\_patch} (baseline): no intervention.
\item \textbf{patch\_A} (full replacement):
$\hh_i^{(\layer^\star)} \leftarrow \hh_j^{(\layer^\star)}$ from the
paired CS donor.
\item \textbf{patch\_B} (mean direction):
$\hh_i^{(\layer^\star)} \leftarrow \hh_i^{(\layer^\star)} +
\overline{\hh_{j}^{(\layer^\star)} - \hh_{i}^{(\layer^\star)}}$
averaged over all pairs (steering-vector analogue).
\item \textbf{ctrl\_rand}: replace with a random non-donor non-self item's hidden state.
\item \textbf{ctrl\_self}: replace with the item's own hidden state (sanity).
\end{itemize}

\paragraph{Results.}
\Cref{tab:patching} reports per-condition accuracy with paired
item-level $95\%$ bootstrap CIs.

\begin{table}[!htbp]
\centering\small
\setlength{\tabcolsep}{4pt}
\begin{tabular}{@{}lccc@{}}
\toprule
Condition & Acc & $95\%$ CI & $n_{\text{correct}}/n$ \\
\midrule
no\_patch (baseline)         & $0.000$ & $[.000, .000]$ & $0/42$ \\
ctrl\_self (sanity)          & $0.000$ & $[.000, .000]$ & $0/42$ \\
patch\_B (mean direction)    & $0.190$ & $[.095, .310]$ & $8/42$ \\
ctrl\_rand (random hidden)   & $0.310$ & $[.167, .452]$ & $13/42$ \\
\textbf{patch\_A (full replace)} & $\mathbf{0.571}$ & $[.429, .714]$ & $24/42$ \\
\bottomrule
\end{tabular}
\caption{Activation patching at Qwen2.5-7B-Instruct layer~$27$ on
the $N{=}42$ anti-CS items where ordinary Yes/No is wrong. The
\textbf{ctrl\_self} sanity matches the baseline at $0$. Full-state
replacement (\textbf{patch\_A}) recovers the correct Yes/No on
$57.1\%$ of items, well above the random-hidden control
(\textbf{ctrl\_rand} $=31.0\%$). The mean-direction analogue
(\textbf{patch\_B} $=19.0\%$) is markedly weaker, indicating the
effective information is item-specific and not fully captured by a
single mean direction.}
\label{tab:patching}
\end{table}

\paragraph{Interpretation.}
ctrl\_self verifies that the hook itself does not perturb the output.
The $0.310$ floor under ctrl\_rand reflects a mild output prior toward
``Yes'' when the layer-$27$ representation is scrambled, so the
meaningful comparison is \textbf{patch\_A vs.\ ctrl\_rand}: replacing
with the matched commonsense donor produces a further
$+26.1$ percentage-point lift over a same-shape random injection. This
goes beyond the linear-readout sufficiency tested by the scalar swap:
the CS-correct last-token residual is \emph{causally sufficient} to
flip the ordinary Yes/No on $>$half of the originally-wrong items,
while being insufficient on the rest. The mean-direction (steering)
analogue captures only a small fraction of this effect, consistent with
item-specific residual content carrying information beyond a single
shared direction.

\section{Forced-choice surface-form robustness}
\label{app:surface}

\paragraph{Setup.}
We re-ask the same $N{=}80$ anti-CS items used in the main scaling
sweep on Qwen2.5-7B-Instruct under four forced-choice surface forms
that share the same prompt body and differ only in the answer slot:

\begin{itemize}\setlength{\itemsep}{0pt}\setlength{\topsep}{2pt}
\item \textbf{Yes/No}: ``\ldots\,does $c$ cause $e$?  Answer (Yes or No):''
\item \textbf{True/False}: ``\ldots\,does $c$ cause $e$?  Answer (True or False):''
\item \textbf{Chinese (\textit{Shi}/\textit{Fou})}: a Chinese rendering of the same forced choice (binary tokens for ``yes''/``no'').
\item \textbf{A/B reframe}: ``\,Which is more plausible \ldots\,A.\ Yes, $c$ causes $e$.\ \ B.\ No, $c$ does not cause $e$. Answer:''
\end{itemize}

In every form, the anti-CS-correct token is the first option
(Yes / True / \textit{Shi} / A).

\paragraph{Results.}
\Cref{tab:surface} reports output accuracy, mean probability on the
correct token, and the gap $\Delta$ relative to the layer-$8$ probe
(probe \Acc{} $0.97$). Across all four forms, output accuracy stays
below $0.4$ and $\Delta \ge +0.595$.

\begin{table}[!htbp]
\centering\scriptsize
\setlength{\tabcolsep}{2.5pt}
\begin{tabular}{@{}lcccc@{}}
\toprule
Surface form & Acc & $95\%$ CI & Mean $p(\text{correct})$ & $\Delta$ vs.\ probe \\
\midrule
Yes/No                  & $0.188$ & $[.100, .263]$ & $0.193$ & $+0.783$ \\
True/False              & $0.325$ & $[.225, .425]$ & $0.331$ & $+0.645$ \\
Chinese (\textit{Shi}/\textit{Fou}) & $0.288$ & $[.188, .388]$ & $0.376$ & $+0.683$ \\
A/B (Yes/No reframed)   & $0.375$ & $[.275, .475]$ & $0.380$ & $+0.595$ \\
\bottomrule
\end{tabular}
\caption{Forced-choice surface-form robustness on
Qwen2.5-7B-Instruct ($N{=}80$ anti-CS items, $320$ forward passes
total). All four surface forms keep output accuracy ${<}0.4$ and
preserve $\Delta\!\ge\!+0.595$ relative to the layer-$8$ probe
(\Acc{} $0.97$). True/False is the most accurate surface form
($0.325$) but still leaves a gap above $+0.6$. The gap is therefore
not specific to the Yes/No surface form, including across a
non-English (Chinese) rendering and an A/B reframe of the same
choice.}
\label{tab:surface}
\end{table}

\begin{figure}[!htbp]
\centering
\includegraphics[width=0.95\columnwidth]{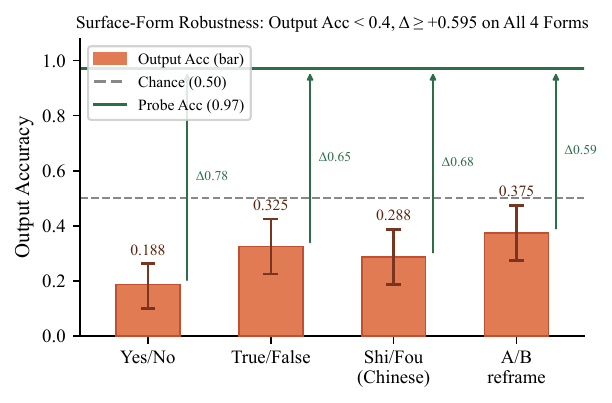}
\caption{Surface-form robustness across four forced-choice answer slots
(Qwen2.5-7B-Instruct, $N{=}80$ anti-CS items).
Bars show output accuracy with 95\% bootstrap CI; the solid green line marks
probe accuracy ($0.97$); the dashed line marks chance ($0.50$).
Green arrows indicate the readout--output gap $\Delta$ for each form.
All four surface forms keep output accuracy below $0.40$ and
$\Delta \ge +0.595$, confirming the gap is not an artefact of the
Yes/No surface form.}
\label{fig:surface_robustness}
\end{figure}

\end{document}